\documentclass{article}

\usepackage{fullpage}
\usepackage{times}
\usepackage{fancyhdr,graphicx,amsmath,amssymb}
\usepackage[ruled,vlined]{algorithm2e}
\include{pythonlisting}

\usepackage{smile}
\usepackage{amsmath}
\usepackage{amssymb}
\usepackage{amsthm}
\usepackage{bbm}
\usepackage{bm}
\usepackage{color}
\usepackage{dirtytalk}
\usepackage{dsfont}
\usepackage{enumerate}
\usepackage{graphicx}
\usepackage{mathtools}
\usepackage{subfigure}
\usepackage{comment}
\usepackage[usenames,dvipsnames]{xcolor}
\usepackage{wrapfig}

\newtheorem{Definition}{Definition}

\newtheorem{Lemma}{Lemma}

\usepackage[final]{neurips_2019}
\usepackage[utf8]{inputenc} % allow utf-8 input
\usepackage[T1]{fontenc}    % use 8-bit T1 fonts
\usepackage{hyperref}       % hyperlinks
\usepackage{url}            % simple URL typesetting
\usepackage{booktabs}       % professional-quality tables
\usepackage{amsfonts}       % blackboard math symbols
\usepackage{nicefrac}       % compact symbols for 1/2, etc.
\usepackage{microtype}      % microtypography
\hypersetup{
	pdftex,
	pdffitwindow=true,
	pdfstartview={FitH},
	pdfnewwindow=true,
	colorlinks,
	linktocpage=true,
	linkcolor=Green,
	urlcolor=Green,
	citecolor=Green
}

\title{Bootstrapping Upper Confidence Bound}

\author{%
 Botao Hao\\
 Purdue University\\
  \texttt{haobotao000@gmail.com} \\
   \And
   Yasin Abbasi-Yadkori \\
  VinAI\\
   \texttt{yasin.abbasi@gmail.com} \\
    \And
  Zheng Wen \\
  Deepmind\\
   \texttt{zhengwen@google.com} \\
    \And
   Guang Cheng\\
  Purdue University\\
   \texttt{chengg@purdue.edu} \\
}

\begin{document}

\maketitle

\begin{abstract}
Upper Confidence Bound (UCB) method is arguably the most celebrated one used in online decision making with partial information feedback. Existing techniques for constructing confidence bounds are typically built upon various concentration inequalities, which thus lead to over-exploration. In this paper, we propose a non-parametric and data-dependent UCB algorithm based on the multiplier bootstrap. To improve its finite sample performance, we further incorporate second-order correction into the above construction. In theory, we derive both problem-dependent and problem-independent regret bounds for multi-armed bandits with symmetric rewards under a much weaker tail assumption than the standard sub-Gaussianity. Numerical results demonstrate significant regret reductions by our method, in comparison with several baselines in a range of multi-armed and linear bandit problems.  
\end{abstract}

%%%%%%%%%%%%%%%%%%%%%%%%%%%%%%%%%%%%%%%
\section{Introduction}\label{sec:intro}
%%%%%%%%%%%%%%%%%%%%%%%%%%%%%%%%%%%%%%%

In artificial intelligence, learning to make decisions online plays a critical role in many fields, such as personalized news recommendation \citep{Li:2010:CAP:1772690.1772758}, robotics \citep{kober2013reinforcement} and the game of Go \citep{silver2016mastering}. To learn to make optimal decisions as soon as possible, the decision-makers must carefully design an algorithm to balance the trade-off between the exploration and exploitation \citep{sutton2018reinforcement, lattimore2018bandit}. Over-exploration could be expensive and unethical in practice, e.g., medical decision making \citep{bastani2015online, bastani2017mostly,bird2016exploring}.  On the other hand, insufficient exploration tends to make an algorithm stuck at a sub-optimal solution. The delicate design of exploration methods stands in the heart of online learning and decision making.

Upper Confidence Bound (UCB) \citep{auer2002using, auer2002finite, dani2008stochastic, li2010contextual, abbasi2011improved} is a class of highly effective algorithms in dealing with the exploration-exploitation trade-off in bandits and reinforcement learning.  
The tightness of confidence bound, as is known, is the key ingredient to achieve the optimal degree of explorations. To the best of our knowledge, nearly all the existing works construct confidence bounds based on various concentration inequalities, e.g. Hoeffding-type \citep{auer2002finite}, empirical Bernstein type \citep{mnih2008empirical} or self-normalized type \citep{abbasi2011improved}. Those concentration-based confidence bounds, however, are typically conservative since they are \emph{data-independent.} Concentration inequalities only exploit tail information, e.g., bounded or sub-Gaussian, rather than the whole distribution knowledge. In general, the loose constant factor may result in confidence bounds that are too wide to be informative \citep{russo2014learning}.

In this paper, we propose a non-parametric and data-dependent UCB algorithm based on the multiplier bootstrap \citep{rubin1981bayesian, wu1986jackknife,  arlot2010some,chernozhukov2014gaussian, spokoiny2015bootstrap}, called bootstrapped UCB. The principle is to use the multiplier bootstrapped quantile as the confidence bound to enforce the exploration. Inspired by recent advances on non-asymptotic guarantee and non-asymptotic inference such as \citep{arlot2010some, chernozhukov2014gaussian, spokoiny2015bootstrap, yang2017non}, we develop an explicit second-order correction for the multiplier bootstrapped quantile that ensures the non-asymptotic validity. Our algorithm is easy to implement and has the potential to be generalized to more complicated models such as structured contextual bandits. 

In theory, we develop both problem-dependent and problem-independent regret bounds for multi-armed bandits with symmetric rewards under a much weaker tail assumption, i.e., sub-Weibull distribution, than the classical sub-Gaussianity. In this case, it is proven that the mean estimator can still achieve the same problem-independent regret bound as the one under the sub-Gaussian assumption. Note that our result does not rely on other sophisticated approaches such as median-of-means or Catoni's M-estimator in \citep{bubeck2013bandits}. A key technical tool we propose is a new concentration inequality for the sum of sub-Weibull random variables. Empirically, we evaluate our method in several multi-armed and linear bandit models. When the exact posterior is unavailable or the noise variance is mis-specified, the bootstrapped UCB demonstrates superior performance over variants of Thompson sampling and concentration-based UCB due to its non-parametric and data-dependent nature. 

Recently, an increasing number of works \citep{elmachtoub2017practical, osband2016deep, tang2015personalized,eckles2014thompson} study bootstrap methods for multi-armed and contextual bandits as an alternative to Thompson sampling. Most treat the bootstrap just as a way to randomize historical data (without any theoretical guarantee). One exception is \citep{kveton2018garbage} who derive a regret bound for Bernoulli bandit by adding pseudo observations. However, their method cannot be easily extended to unbounded cases, and their analyses heavily limit to the Bernoulli assumption. In contrast, our method applies to a broader class of bandit models with rigorous regret analysis. 

The rest of the paper is organized as follows. Section \ref{sec:boostrap_UCB} introduces the basic setup and our bootstrapped UCB algorithm. Section \ref{sec:regret} provides the regret analysis and Section \ref{sec:exp} conducts several experiments.

\paragraph{Notations.}  Throughout the paper, we denote $\mathbb P_{\bw}(\cdot), \mathbb E_{\bw}(\cdot)$ as the probability and expectation operator with respect to the distribution of the vector $\bw$ only, conditioning on other random variables. We use similar notations for $\mathbb P_{\by}(\cdot)$, $\mathbb E_{\by}(\cdot)$ with respect to $\by$ only.  $[n]$ means the set $\{1,2,\ldots, n\}$. We denote boldface lower letters (e.g. $\bx$, $\by$) as a vector. For a set $\cE$, we define its complement as $\cE^c$.

%%%%%%%%%%%%%%%%%%%%%%%%%%%%
\section{Bootstrapped UCB}\label{sec:boostrap_UCB}
%%%%%%%%%%%%%%%%%%%%%%%%%%%%

\paragraph{Problem setup.} As a fruit fly, we illustrate our idea on the stochastic multi-armed bandit problem \citep{lai1985asymptotically, lattimore2018bandit}. In detail, the decision-makers interact with an environment for $T$ rounds. In round $t\in[T]$, the decision-makers pull an arm $I_t\in[K]$ and observes its reward $y_{I_t}$ which is drawn from a distribution associated with the arm $I_t$, denoted by $P_{I_t}$ with an unknown mean $\mu_{I_t}$. Without loss of generality, we assume arm $1$ is the optimal arm, that is, $\mu_1 = \max_{k\in[K]} \mu_k$. In multi-armed bandit problems, the objective is to minimize the expected cumulative regret,
defined as,
\begin{equation}\label{def:regret}
    R(T)=T\mu_1 - \mathbb E\Big[\sum_{t=1}^Ty_t\Big] = \sum_{k=2}^K \Delta_k\mathbb E\Big[\sum_{t=1}^T \mathbf{I}\{I_t = k\}\Big],
\end{equation}
where $\Delta_k = \mu_1-\mu_k$ is the sub-optimality gap for arm $k$, and $\mathbf{I}\{\cdot\}$ is an indicator function. Here, the second equality is from the regret decomposition Lemma (Lemma 4.5 in \citep{lattimore2018bandit}). We call an upper bound of $R(T)$ problem-independent if the bound only depends on the distributional assumption and not on the specific bandit problem, say the gap $\Delta_k$. 

\paragraph{Upper Confidence Bound.} The upper confidence bound (UCB) algorithm \citep{auer2002finite} is based on the principle of optimism in the face of uncertainty. The key idea is to act as if the environment (parameterized by $\mu_k$ in multi-armed bandits) is as nice as plausibly possible. Concretely, a plausible environment refers to an upper confidence bound $\cG(\by_n, 1-\alpha)$ for the true mean $\mu$, of the form
\begin{equation}\label{def:confidence_interval}
    \cG(\by_n, 1-\alpha) = \big\{x\in \mathbb R , x-\bar{y}_n \leq  h_{\alpha}(\by_n)\big\},
\end{equation}
where $\by_n = (y_1, \ldots, y_n)^{\top}$ is the sample vector, $\bar{y}_n$ is the empirical mean, $\alpha \in(0, 1)$ is the confidence level, and $h_{\alpha}: \mathbb R^n \to \mathbb R^+$ is a threshold that could be either data-dependent or data-independent.

\begin{definition}\label{def:CI}
We define $\cG(\by_n, 1-\alpha)$ as a non-asymptotic upper confidence bound if for \emph{any sample size $n\geq 1$}, the following inequality holds
\begin{equation}
    \mathbb P\Big(\mu\in \cG(\by_n, 1-\alpha)\Big) \geq 1-\alpha.
\end{equation}
\end{definition}
In bandit problems, a non-asymptotic control on the confidence level is more commonly used. This is rather different from the asymptotic validity of confidence bound in statistics literature \citep{casella2002statistical}. 

A generic UCB algorithm will select the action based on its UCB index $\bar{y}_{n} + h_{\alpha}(\by_{n})$ for different arms.
 As is well known, the sharper the threshold is, the better exploration and exploitation trade-off one can achieve \citep{lattimore2018bandit}. By the definition of quantile, the sharpest threshold in \eqref{def:confidence_interval} is the $(1-\alpha)$-quantile of the distribution of $\bar{y}_n - \mu$. However, this quantile relies on the knowledge of the exact reward distribution and is therefore itself unknown. To evaluate this value, we construct a data-dependent confidence bound based on the multiplier bootstrap.

\subsection{Confidence Bound Based on Multiplier Bootstrap}

\paragraph{Multiplier Bootstrap.}

Multiplier bootstrap is a fast and easy-to-implement alternative to the standard bootstrap, and has been successfully applied in various statistical contexts \citep{arlot2010some,chernozhukov2014gaussian, spokoiny2015bootstrap}. Its goal is to approximate the distribution of the target statistic by
reweighing its summands with random multipliers independent of the data. For instance, in a mean estimation problem, we define a multiplier bootstrapped estimator as $n^{-1}\sum_{i=1}^n w_{i}(y_{i} - \bar{y}_n) = n^{-1}\sum_{i=1}^n(w_{i} - \bar{w}_n)y_{i},$
where $\{w_{i}\}_{i=1}^n$ are some random variables independent of $\by_n$, called bootstrap weights. Some classical weights are as follows:
\begin{itemize}
    \item \emph{Efron's bootstrap weights.} $(w_{1}, \ldots, w_{n})$ is a multinomial random vector with parameters $(n; n^{-1}, \ldots, n^{-1})$. This is the standard nonparameteric bootstrap \citep{efron1982jackknife}.
    \item \emph{Gaussian weights.} $w_{i}$'s are i.i.d standard Gaussian random variables. This is closely related to Gaussian approximation in statistics \citep{chernozhukov2014gaussian}.
    \item \emph{Rademacher weights.} $w_{i}$'s are i.i.d Rademacher variables. This is closely related to symmetrization in learning theory.
\end{itemize}

The bootstrap principle suggests that the $(1-\alpha)$-quantile of the distribution of $  n^{-1} \sum_{i=1}^n w_{i}(y_{i} - \bar{y}_n)$ conditionally on $\by_n$ could be used to approximate the $(1-\alpha)$-quantile of the distribution of $\bar{y}_n-\mu$. As the first building block, the multiplier bootstrapped quantile is defined as,
\begin{equation}\label{def:quantile}
    q_{\alpha}(\by_n-\bar{y}_n):=\inf\Big\{x\in\mathbb R |\mathbb P_{\bw}\Big(\frac{1}{n}\sum_{i=1}^n w_{i}(y_{i}-\bar{y}_n)>x\Big)\leq \alpha\Big\}.
\end{equation}
The question is whether $ q_{\alpha}(\by_n-\bar{y}_n)$ is a valid threshold for any sample size $n\geq 1$.

\subsection{Second-order Correction}

Most statistical theories guarantee the {\em asymptotic validity} of $q_{\alpha}(\by_n-\bar{y}_n)$ by the multiplier central limit theorem \citep{van2000asymptotic}. However, we show that such a claim is valid \emph{non-asymptotically} at the cost of adding a second-order correction. Next theorem rigorously characterizes this phenomenon under a symmetric assumption on the reward. Moreover, in Section \ref{subsec:linear_regret} in the supplement, we show that without the second-order correction, a naive bootstrapped UCB will result in linear regret.

\begin{theorem}[Non-asymptotic Second-order Correction]\label{thm:non_asy_bootstrap}
Suppose $\{y_i\}_{i=1}^n$ are i.i.d symmetric random variables with respect to its mean $\mu$, and the bootstrap weights $\{w_i\}_{i=1}^n$ are i.i.d Rademacher random variables. 
For two arbitrary parameters $\alpha, \delta\in(0,1)$, the following inequality holds for any sample size $n\ge 1$,
\begin{equation}\label{eqn:non_asy}
    \mathbb P_{\by} \Big(\bar{y}_n -\mu > \underbrace{q_{\alpha(1-\delta)}(\by_n-\bar{y}_n) + \sqrt{\frac{\log (2/\alpha\delta)}{n}}\varphi(\by_n)}_{\text{bootstrapped threshold}}\Big) \leq 2\alpha,
\end{equation}
where $\varphi(\by_n)$ is a non-negative function satisfying $\mathbb P_{\by}(|\bar{y}_n-\mu| \geq \varphi(\by_n)) \leq \alpha.$
\end{theorem}
The detailed proof is deferred to Section \ref{proof:non_asy} in the supplement. In \eqref{eqn:non_asy}, the bootstrapped threshold may be interpreted as a main term, i.e., $q_{\alpha(1-\delta)}(\by_n-\bar{y}_n) $ (at a shrunk confidence level), plus a second-order correction term, i.e., $(\log(2/\alpha\delta)/n)^{1/2}\varphi(\by_n)$. The latter is added to guarantee the non-asymptotic validity of the bootstrapped threshold. In the above, $\varphi(\by_n)$ could be any preliminary upper bound on $\bar{y}_n -\mu$. Hence, Theorem \ref{thm:non_asy_bootstrap} transforms a possibly coarse prior bound $\varphi(\by_n)$ on quantiles into a more accurate version that is based on a main term estimated by multiplier bootstrap plus a second-order correction term based
on $\varphi(\by_n)$ multiplied by a $\cO(n^{-1/2})$ factor.

\begin{remark}[Choice of $\varphi(\by_n)$]
If $\{y_{i}\}_{i=1}^n$ are independent 1-sub-Gaussian random variables, a natural choice of $\varphi(\by_n)$ is $(2\log (1/\alpha)/n)^{1/2}$ by Hoeffding's inequality (Lemma \ref{lemma:hoeffding}). Plugging it into \eqref{eqn:non_asy} and letting $\delta = 1/2$, the bootstrapped threshold in \eqref{eqn:non_asy} becomes
\begin{equation}\label{def:boostrap_CI}
   \underbrace{q_{\alpha/4}(\by_n-\bar{y}_n)}_{\text{main term}} + \underbrace{\frac{2\log (8/\alpha)}{n}}_{\text{second order correction}}.
\end{equation}
Lemma \ref{lemma:quantile_bound} in the supplement shows that the main term is of order at least $\cO(n^{-1/2})$ as $n$ grows, which implies the second order correction is just a remainder term. We emphasize that the reminder term is obviously not sharp and will be sharpened as a future work. 
\end{remark}

\begin{remark}
Existing works on UCB-type algorithms typically utilized various concentration inequalities, e.g. Hoeffding's inequality \citep{auer2002finite} or empirical Bernstein's inequality \citep{mnih2008empirical}, to find a valid threshold $h_{\alpha}(\by_n)$. However, they are not data-dependent and only use the tail information, rather than fully exploit the whole distribution knowledge. This is typically conservative, and leads to over-exploration. 
\end{remark}

\begin{remark}
Empirical KL-UCB \citep{cappe2013kullback} used empirical likelihood to build confidence intervals for general distributions that have support in $[0, 1]$. Although empirical KL-UCB is also data-dependent, our proposed method is from a very different non-parametric perspective and uses different tools by bootstrap.  In practice, resampling tends to be more efficient computationally, without solving a convex optimization each round like empirical KL-UCB. Moreover, our method can work with unbounded rewards and we believe it is easier to generalize to structured bandits, e.g. linear bandit.
\end{remark}

\begin{figure}[h]
	\centering
\includegraphics[scale=0.45]{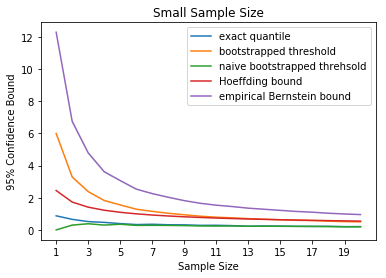}
		\includegraphics[scale=0.45]{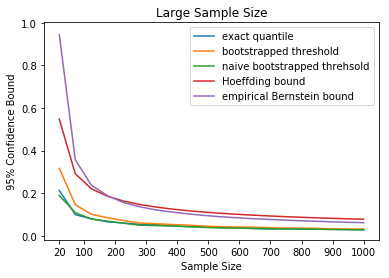}
			\vspace{-0.13in}
\caption{95\% confidence bound of the sample mean.}\label{plt:quantile}
\end{figure}

In Figure \ref{plt:quantile}, we compare different approaches to calculate 95\% confidence bound for the population mean based on samples from a truncated-normal distribution. When the sample size is extremely small $(\le 10)$, the naive bootstrap (without any correction) cannot output a valid threshold since the bootstrapped quantile is smaller than the true 95\% quantile. This confirms the necessity of the second-order correction. When the sample size increases, our bootstrapped threshold converges to the truth rapidly. This confirms the correction term is just a small remainder term. Additionally, the bootstrapped threshold is shown to be sharper than Hoeffding's bound and empirical Bernstein bound when sample size is large (see the right panel of Figure \ref{plt:quantile}).

\subsection{Main Algorithm: Bootstrapped UCB}

Based on the above theoretical findings, we conclude that bootstrapped UCB will select the arm according to its UCB index defined as below:
\begin{equation}\label{eqn:bootstrapped_UCB}
    \text{UCB}_k(t) = \bar{y}_{n_{k,t}} + q_{\alpha(1-\delta)}(\by_{n_{k,t}}-\bar{y}_{n_{k,t}}) + \sqrt{\frac{\log (2/\alpha\delta)}{n_{k,t}}}\varphi(\by_{n_{k,t})} \;,
\end{equation}
where $n_{k,t}$ is the number of pulls for arm $k$ until time $t$. Practically, we may use Monte Carlo quantile approximation to get an approximated bootstrapped quantile $\tilde{q}_{\alpha(1-\delta)}(\by_{n_{k,t}}-\bar{y}_{n_{k,t}}, \bw^B)$ and corresponding theorem for the control of the approximation of the bootstrapped quantile is also derived (see Section \ref{sec:monte_carlo_approximation} in the supplement for details). The algorithm is summarized in Algorithm \ref{alg:bucb}. The computational complexity at step $t$ is $\tilde{\cO}(Bt) \leq \tilde{\cO}(BT)$. Comparing with vanilla UCB, the extra $Bt$ is due to resampling. In practice, the choice of $B$ is seldom treated as a tuning parameter, but usually determined by the available computational resource.
\begin{algorithm}[H]
\SetAlgoLined
\KwIn{the number of bootstrap repetitions $B$, hyper-parameter $\delta$.}
 \For{$t=1$ to $K$}{
  Pull each arm once to initialize the algorithm.\;
  }
  \For{$t = K+1$ to $T$}{
  Set confidence level $\alpha = 1/(t+1)$.
  
  Calculate the boostrapped quantile $\tilde{q}_{\alpha(1-\delta)}(\by_{n_{k,t}}-\bar{y}_{n_{k,t}}, \bw^B)$. \;
  
  Pull the arm $$I_t = \argmax_{k\in[K]}(\bar{y}_{n_{k,t}} + \tilde{q}_{\alpha(1-\delta)}(\by_{n_{k,t}}-\bar{y}_{n_{k,t}}, \bw^B) + (\log (2/\alpha\delta)/n_{k,t})^{1/2}\varphi(\by_{n_{k,t}})).$$\;
  
  Receive reward $y_{I_t}$.\;
  }
 \caption{Bootstrapped UCB}\label{alg:bucb}
\end{algorithm}

%%%%%%%%%%%%%%%%%%%%%%%%%%%%
\section{Regret Analysis}\label{sec:regret}
%%%%%%%%%%%%%%%%%%%%%%%%%%%%
In Section \ref{subsec:log_regret}, we derive regret bounds for bootstrapped UCB. Moreover, we show that naive bootstrapped UCB will result in linear regret in some cases in Section \ref{subsec:linear_regret} in the supplement.

%%%%%%%%%%%%%%%%%%%%%%%%%%%%%%%%%%%%%%%%%%%%%%%%%%%%%%%
\subsection{Regret Bound for Bootstrapped UCB}\label{subsec:log_regret}
%%%%%%%%%%%%%%%%%%%%%%%%%%%%%%%%%%%%%%%%%%%%%%%%%%%%%%%

For multi-armed bandit problems, most literature \citep{lattimore2018bandit} consider sub-Gaussian rewards. In this work, we move beyond sub-Gaussianity and consider the reward under a much weaker tail assumption, so-called sub-Weibull distribution. As shown in \citep{kuchibhotla2018moving,vladimirova2019sub}, it is characterized by the right tail of the Weibull distribution and generalizes sub-Gaussian and sub-exponential distributions. 
 \begin{Definition}[(Sub-Weibull Distribution)]
	We define $y$ as a sub-Weibull random variable if it has a bounded $\psi_{\beta}$-norm. The $\psi_{\beta}$-norm of $y$ for any $\beta>0$ is defined as
	\begin{eqnarray*}
		\|y\|_{\psi_{\beta}}:=\inf\Big\{C\in(0, \infty): ~ \mathbb E[\exp(|y|^{\beta}/C^{\beta})]\leq 2\Big\}.
	\end{eqnarray*}
\end{Definition}
Particularly, when $\beta$ = 1 or 2, sub-Weibull random variables reduce to sub-exponential or sub-Gaussian random variables, respectively. It is obvious that the smaller $\beta$ is, the heavier tail the random variable has. Next theorem provides a corresponding concentration inequality for the sum of independent sub-Weibull random variables. 
\begin{theorem}[Concentration Inequality for Sub-Weibull Distribution]\label{thm:orlicz_concentra}
Suppose  $\{y_i\}_{i=1}^n$ are independent sub-Weibull random variables with $\|y_i\|_{\psi_{\beta}}\leq \sigma$. Then there exists an absolute constant $C_{\beta}$ only depending on $\beta$ such that for any $\ba=(a_1,\ldots,a_n)\in \mathbb R^n$ and $0<\alpha<1/e^2$,
	\begin{eqnarray*}
		\Big|\sum_{i=1}^na_iy_i-\mathbb E(\sum_{i=1}^na_iy_i)\Big|\leq C_{\beta}\sigma\Big(\|\ba\|_2(\log\alpha^{-1})^{1/2}+\|\ba\|_{\infty}(\log \alpha^{-1})^{1/\beta}\Big)
	\end{eqnarray*}
	with probability at least $1-\alpha$.
\end{theorem}
The proof relies on a precise characterization of $p$-th moment of a Weibull random variable and standard symmetrization arguments. Details are deferred to Section \ref{proof:concentration} in the supplement. This theorem generalizes the Hoeffding-type concentration inequalities for sub-Gaussian random variables (see, e.g. Proposition 5.10 in \cite{RV12}), and Bernstein-type concentration inequalities for sub-exponential random variables (see, e.g. Proposition 5.16 in \cite{RV12}) up to some constants.

In Theorem \ref{thm:regret_problem_dependent}, we provide both problem-dependent and problem-independent regret bounds.

\begin{theorem}\label{thm:regret_problem_dependent}
Consider a stochastic $K$-armed sub-Weibull bandit, where the noise follows a symmetric sub-Weibull distribution with its $\psi_{\beta}$-norm upper bounded by $\sigma$. Denote $n_{k,t}$ as the number of pulls for arm $k$ until time $t$. We choose $\varphi$ according to Theorem \ref{thm:orlicz_concentra} as follows
\begin{equation}\label{eqn:varphi}
    \varphi(\by_{n_{k,t}}) = C_{\beta}\sigma \Big(\sqrt{\frac{\log 1/\alpha}{n_{k,t}}} + \frac{(\log 2/\alpha)^{1/\beta}}{n_{k,t}}\Big),
\end{equation}
 and let the confidence level $\alpha = 1/T^2$. For any round $T$, the problem-dependent regret of bootstrapped UCB is upper bounded by
\begin{equation}\label{eqn:problem_dep}
    R(T) \leq \sum_{k:\Delta_k>0} 128C_{\beta}^2\sigma^2\frac{\log T}{\Delta_k} + 2^{3+1/\beta}C_{\beta}\sigma K(\log T)^{1/\beta} + 4\sum_{k=2}^K\Delta_k,
\end{equation}
where $C_{\beta}$ is some absolute constant from Theorem \ref{thm:orlicz_concentra}, and $\Delta_k$ is the sub-optimality gap. Moreover, if the round $T\geq 2^{2/\beta-3}K(\log T)^{2/\beta-1}$, the problem-independent regret of bootstrapped UCB is upper bounded by
\begin{eqnarray}
    R(T)\leq 32\sqrt{2}C_{\beta}\sigma\sqrt{TK\log T} +4K\mu_1^*.
\end{eqnarray}

\end{theorem}
The main proof structure follows the standard analysis of UCB \citep{lattimore2018bandit} and relies on a sharp upper bound for the (data-dependent) bootstrapped quantile term by Theorem \ref{thm:orlicz_concentra}. Details are deferred to Section \ref{proof:main} in the supplement. When $\beta\geq1$, \eqref{eqn:problem_dep} provides a logarithm regret that matches the state-of-art result \citep{lattimore2018bandit}. When $\beta<1$, we have a non-negligible term $(\log T)^{1/\beta}$ that is the price paid for heavy-tailedness. However, this term does not depend on the gap $\Delta_k$. Therefore, we have an optimal problem-independent regret bound. 

\begin{remark}
The choice of $\alpha = 1/T^2$ led to an easy analysis. Using similar techniques in Chapter 8.2 of \citep{lattimore2018bandit}, we can achieve a similar regret bound by setting $\alpha_t = 1/(t\log^{\tau}(t))$ for any $\tau>0$.
\end{remark}
\begin{remark}
\citep{bubeck2013bandits} consider bandit with heavy-tail (moment of order $(1+\varepsilon)$) based on a median-of-means estimator. As mentioned in \cite{chen2019robust}, there are two disadvantages for median-of-means approach: (a) it involves an additional tuning parameter; (b) it is numerically unstable for small sample size. In contrast, we identify a class of heavy-tailed bandits (sub-Weibull bandit) where mean estimators can still achieve regret bounds of the same order as those under sub-Gaussian reward distributions. The reason is that although sub-Weibull r.v. has heavier tail than sub-Gaussian r.v., its tail still has an exponential-like decay.
\end{remark}

%%%%%%%%%%%%%%%%%%%%%
\section{Experiments}\label{sec:exp}
%%%%%%%%%%%%%%%%%%%%%
 In Section \ref{subsec:sim_multi_bandit}, we consider multi-armed bandits with both symmetric and asymmetric rewards.  In Section \ref{subsec:sim_linear_bandit}, we extend our method to linear bandits. %The code is written in python and implemented on an Intel XeonE5 processor with 64 GB of RAM. 
 Implementation details and some additional experimental results are deferred to Section \ref{sec:addition_simu} in the supplement.

%%%%%%%%%%%%%%%%%%%%%%%%%%
\subsection{Multi-armed Bandit}\label{subsec:sim_multi_bandit}
%%%%%%%%%%%%%%%%%%%%%%%%%%
In this section, we compare bootstrapped UCB (Algorithm \ref{alg:bucb}) with three baselines: Upper Confidence Bound based on concentration inequalities (Vanilla UCB), Thompson sampling with normal Jeffery prior \citep{korda2013thompson} (Jeffery-TS) and Thompson sampling with Beta prior \citep{agrawal2013further} (Bernoulli-TS). For bounded rewards, we also compare with Giro \citep{kveton2018garbage}\footnote{We have implemented Giro in the unbounded reward case, which could result in linear regret in most cases. See Figure \ref{plt:logis} in the supplement. So, it's unclear what is the best way to add pseudo observations in this case.}, that is a sampling-based exploration method by adding artificial pseudo observations $\{0, 1\}$ to escape from local optima, and empirical KL-UCB \citep{garivier2011kl} using package: \href{https://www.di.ens.fr/~cappe/Code/PymaBandits/}{PymaBandits}. For the preliminary bound $\varphi(\by_n)$, we simply choose the one derived by the concentration inequality. Note that the second-order correction term in \eqref{eqn:non_asy} is conservative. For practitioners, we suggest to set the correction term to be $\varphi(\by_n)/\sqrt{n}$. To be fair, we choose the confidence level $\alpha = 1/(1+t)$ for both UCB1 and bootstrapped UCB, and $\delta = 0.1$ in \eqref{eqn:non_asy}. All algorithms above require knowledge of an upper bound on the noise standard deviation. The number of bootstrap repetitions is $B=200$, and the number of arms is $K=5$.

\begin{figure}[h]
	\centering

		\includegraphics[scale=0.4]{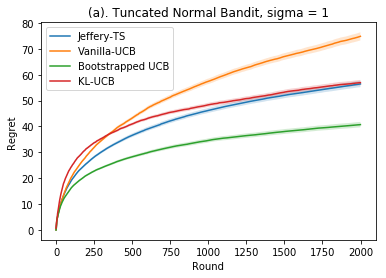}
		\includegraphics[scale=0.4]{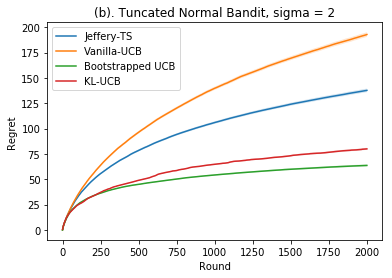}
		\includegraphics[scale=0.4]{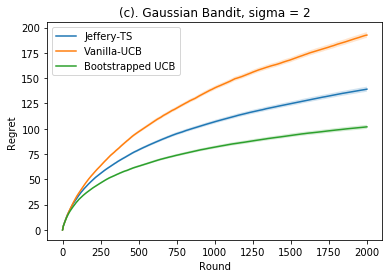}
		\includegraphics[scale=0.4]{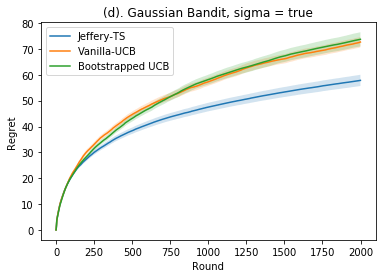}
		\vspace{-0.13in}
\caption{Cumulative regrets for truncated-normal bandit and Gaussian bandit. Sigma is the upper bound on the standard deviation of the noise. The results are averaged over 200 realizations. }\label{plt_unbounded}
\end{figure}

First, we consider symmetric rewards with a mean parameter $\mu_k$ generated from $\text{Uniform}(-1,1)$. The noise follows either truncated-normal distribution within $[-1, 1]$, or standard Gaussian distribution. From Figure \ref{plt_unbounded}, bootstrapped UCB outperforms Jeffery-TS and Vanilla-UCB for truncated-normal bandit and has comparable or sometimes better performance over empirical KL-UCB.   It's obvious that if the reward distribution is exactly Gaussian and the plug-in estimate for the noise standard deviation is the truth, Jeffery-TS should be the best. However, when the posterior (plots (a),(b)) or noise standard derivation (plot (c)) are mis-specified, the performance of TS deteriorates fast. Since (concentration-based) Vanilla UCB only uses the tail information (bounded or sub-Gaussian), it is very conservative and results in bad regret as expected. 

\begin{figure}[h]
	\centering

		\includegraphics[scale=0.4]{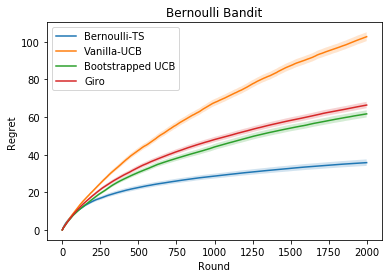}
		\includegraphics[scale=0.4]{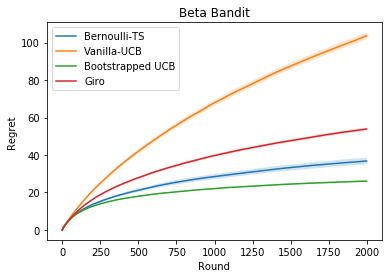}
		\vspace{-0.13in}
\caption{Cumulative regrets for Bernoulli bandit and Beta bandit. The results are averaged over 200 realizations.}\label{plt:bounded}
\end{figure}

Second, we consider asymmetric rewards with a mean parameter $\mu_k$ generated from $\text{Uniform}(0.25, 0.75)$. For Bernoulli bandit, the reward follows $\mathrm{Ber}(\mu_k)$; for Beta bandit, the reward follows \footnote{We adopt the technique in \cite{agrawal2013further} to run Thompson Sampling with $[0, 1]$ rewards. In particular, for any reward $y_t \in [0, 1]$, we draw pseudo reward $\hat{y}_t \sim \mathrm{Ber}(y_t)$, and then use $\hat{y}_t$ instead of $y_t$ in the algorithm.} $\mathrm{Beta}(v \mu_k, v (1 - \mu_k))$ for $v =8$. From Figure \ref{plt:bounded}, bootstrapped UCB outperforms Vanilla UCB and Giro in both cases, and outperforms Bernoulli-TS for Beta bandit. In fact, we are supposed not to beat Bernoulli-TS for Bernoulli bandit since TS fully makes use of the distribution knowledge in this case. One possible explanation is that our method is non-parametric. 

\begin{figure}[h]
	\centering
		\includegraphics[scale=0.4]{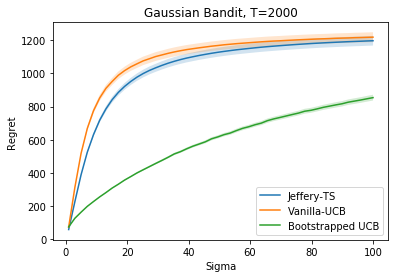}
		\includegraphics[scale=0.4]{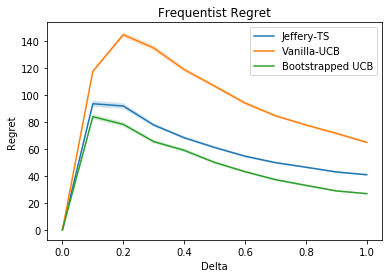}
		\vspace{-0.13in}
\caption{The left panel is the cumulative regret over noise levels while the right panel is the instance-dependent regret of various algorithms as a function of gaps. The results are averaged over 200 realizations.}\label{plt:variance}
\end{figure}

Third, we demonstrate that the robustness of bootstrapped UCB over mis-specifications of the noise standard deviation. In the left panel of Figure \ref{plt:variance}, we consider the cumulative regret at round $T=2000$ of standard Gaussian bandit. As one can see, when we increase the plug-in upper bound of the standard deviation of the noise, bootstrapped UCB is more robust than Bernoulli-TS and Vanilla UCB.

Last, we present a frequentist instance-dependent regret curve for truncated-normal bandit and the experiment set up follows \cite{lattimore2018refining}. We plot cumulative regrets at $T=2000$ of various algorithms with respect to the instance gap $\Delta$ and the mean vector $\mu=(\Delta, 0, 0, 0, 0)$. The results are summarized in the right panel of Figure \ref{plt:variance}.

%%%%%%%%%%%%%%%%%%%%%%%%%%
\subsection{Linear Bandit}\label{subsec:sim_linear_bandit}
\begin{figure}[h]
	\centering
		\includegraphics[scale=0.4]{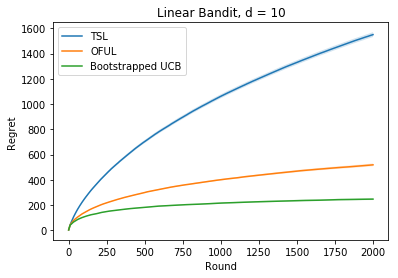}
		\includegraphics[scale=0.4]{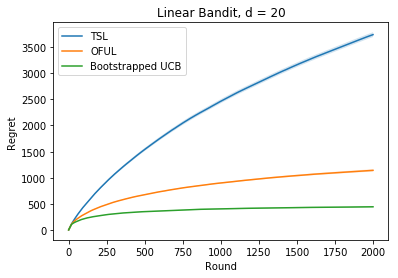}
		\vspace{-0.13in}
\caption{Cumulative regret for linear bandit.}\label{plt:linear_bandit}
\end{figure}

%%%%%%%%%%%%%%%%%%%%%%%%%%
We extend our method to linear bandit case. The basic set up follows the one in \cite{russo2014learning}. In detail, $\btheta^* \in\mathbb R^{d}$ is drawn from a multivariate Gaussian distribution with mean vector $\mu=0$ and covariance matrix $\Sigma = 10I_d$. The noise follows a standard Gaussian distribution. There are $100$ actions with feature vector components drawn uniformly at random from $[-1/\sqrt{10}, 1/\sqrt{10}]$. We consider two state-of-art methods: Thompson sampling for linear bandit \citep{agrawal2013thompson} (TSL) and optimism in the face of uncertainty for linear bandits \citep{abbasi2011improved} (OFUL). 
Following the principle of constructing second-order correction in mean problems (Theorem \ref{thm:non_asy_bootstrap}), we construct the bootstrapped UCB for linear bandit (BUCBL) as follows: At each round $t$, the action is selected as  $\argmax_{\bx}(\bx^{\top}\hat{\btheta}_t + \beta^{\text{BUCBL}}_{t,1-\delta}\|\bx\|_{V_t^{-1}})$, where $\beta^{\text{BUCBL}}_{t,1-\delta} = q_{\alpha}(\hat{\btheta}_t^{(b)}-\hat{\btheta}_t)+ \beta^{\text{OFUL}}_{t,1-\delta, \sigma}/\sqrt{n}$. The formal definition of $\hat{\btheta}_t, \hat{\btheta}_t^{(b)}, \beta^{\text{OFUL}}_{t,1-\delta, \sigma}$ and some basic setups are given in Section \ref{sec:linear_bandit} in the supplement.
To be fair, the confidence level for all methods is set to be $\delta = 1/(1+t)$ and we plug in the true standard deviation of the noise for each method. From Figure \ref{plt:linear_bandit}, we can see that bootstrapped UCB greatly improves the cumulative regret over TSL and OFUL. 

%%%%%%%%%%%%%%%%%%%%%%%%%%%%%%%%%%%
\section{Conclusion}\label{sec:con}
%%%%%%%%%%%%%%%%%%%%%%%%%%%%%%%%%%%%
In this paper, we propose a novel class of non-parametric and data-driven UCB algorithms based on multiplier bootstrap. It is easy to implement and has the potential to be generalized to other complex structured problems. As future works, we will evaluate our idea on other structured contextual bandits and reinforcement learning problems.

\subsubsection*{Acknowledgments}

We thank Tor Lattimore for helpful discussions. Guang Cheng would like to acknowledge support by NSF DMS-1712907, DMS-1811812, DMS-1821183, and Office of Naval Research (ONR N00014-18-2759). In addition, Guang Cheng is a visiting member of Institute for Advanced Study, Princeton (funding provided by Eric and Wendy Schmidt) and visiting Fellow of SAMSI for the Deep Learning Program in the Fall of 2019; he would like to thank both Institutes for their hospitality.

%\bibliographystyle{unsrt}%Used BibTeX style is unsrt
%{\small
%\bibliography{neurips_2019}
%}
\bibliography{neurips_2019}
\bibliographystyle{plainnat}

\clearpage
\onecolumn
\appendix
\begin{center}
    \Large Supplement to ``Bootstrapping Upper Confidence Bound''
\end{center}

In this supplement, we provide linear regret result in Section \ref{subsec:linear_regret}, major proofs in Sections \ref{sec:proof_thm} and \ref{proof:main_lemmas}. Some implementation details are in Sections \ref{sec:monte_carlo_approximation} and \ref{sec:addition_simu}. In the end, we provide several supporting lemmas in Section \ref{sec:supporing_lemma}.

\section{Linear Regret}\label{subsec:linear_regret}

Following the augments in \cite{vaswani2018new, kveton2018garbage}, in this section, we show that UCB with a naive bootstrapped confidence bound will result in linear regret in two-armed Bernoulli bandit. At round $t+1$,  the UCB index without the correction term for arm $k$ can be written as
\begin{equation*}
    \text{UCB}_k(t) = \bar{y}_{n_{k,t}} +q_{\alpha(1-\delta)}(\by_{n_{k,t}}-\bar{y}_{n_{k,t}}).
\end{equation*}
Consider the case where
the first observation on the optimal arm is 0 but on the sub-optimal arm is 1. A key fact is that if the rewards are all zero, no matter how you bootstrap the data, the bootstrapped quantile is always zero. This will make the algorithm stuck into the sub-optimal arm.

\begin{theorem}\label{thm:linear_regret}
Consider a stochastic 2-arm Bernoulli bandit with mean parameter $\mu_1,\mu_2$. The expected regret of the naive bootstrapped UCB can be lower bounded by 
\begin{equation}
    R(T)\geq \Delta_2\Big((1-\mu_1)\mu_2(T-2)+1\Big).
\end{equation}
\end{theorem}

\paragraph{Proof.} Without loss of generality, we assume arm 1 is the optimal arm. Suppose at round $t= 1, 2$, we pull each arm once such that $y_1$ is with arm 1 and $y_2$ is with arm 2. Then we define a bad event as follows:
\begin{equation}
    \cE = \{y_1 = 0, y_2 = 1\}.
\end{equation}
We know that under event $\cE$, the decision-maker will never pull arm 1 any more starting from round $t=3$. This is because if the rewards are all zero, no matter how you bootstrap the data, the bootstrapped quantile is always zero and then makes the decision-maker struck into the sub-optimal arm. Finally, we can lower bound the cumulative regret by,
\begin{eqnarray*}
    R(T) &=& \Delta_2\mathbb E\Big[\sum_{t=1}^T\mathbf{I}\{I_t = 2\}\Big]\\
    &=&\Delta_2\mathbb E\Big[\sum_{t=3}^T\mathbf{I}\{I_t = 2\}|\cE\Big] \mathbb P(\cE) + \Delta_2\mathbb E\Big[\sum_{t=3}^T\mathbf{I}\{I_t = 2\}|\cE^c\Big] \mathbb P(\cE^c) + \Delta_2\\
    &\geq& \Delta_2\mathbb E\Big[\sum_{t=3}^T\mathbf{I}\{I_t = 2\}|\cE\Big] \mathbb P(\cE)+ \Delta_2\\
    &=&\Delta_2 T \mathbb P(y_1 = 0)\mathbb P(y_2 = 1)+ \Delta_2\\
    &=& \Delta_2\Big((1-\mu_1)\mu_2(T-2)+1\Big).
\end{eqnarray*}
This ends the proof. \hfill $\blacksquare$\\

\begin{figure}[h]
	\centering
\includegraphics[scale=0.45]{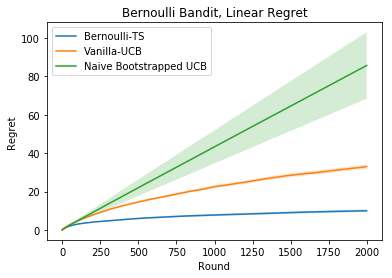}
		\includegraphics[scale=0.45]{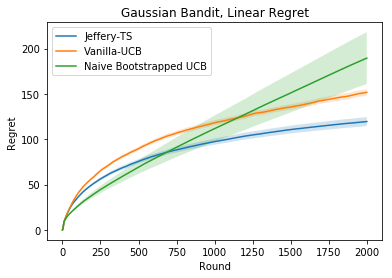}
			\vspace{-0.13in}
\caption{Linear regret of naive bootstrapped UCB on Bernoulli bandit and Gaussian bandit. The result is averaged over 200 realizations.}\label{plt:gs_linear}
\end{figure}

We further demonstrate this phenomenon empirically for both Bernoulli  bandit and Gaussian bandit in Figure \ref{plt:gs_linear}.

\section{Proofs of Main Theorems}\label{sec:proof_thm}
In this section, we provide detailed proofs of Theorems \ref{thm:non_asy_bootstrap}, \ref{thm:orlicz_concentra} and \ref{thm:regret_problem_dependent}.

\subsection{Proof of Theorem \ref{thm:non_asy_bootstrap}}\label{proof:non_asy}

The proof borrows the analysis from \cite{arlot2010some} but with refined analysis and sharp large deviation bound for binomial random variables. 
\paragraph{Step One.} Recall that \eqref{def:quantile} can be seen as the multiplier bootstrapped quantile around its empirical mean. We first takes advantage of the symmetry of each $\by$ around its mean by connecting the true quantile of $\bar{y}_n-\mu$ and the multiplier bootstrapped quantile around the true mean. Define the multiplier bootstrapped quantile around the true mean as 
\begin{equation}\label{def:quantitle_mean}
    q_{\alpha}( \by_{n}-\mu):=\inf\Big\{x\in\mathbb R |\mathbb P_{\bw}\Big(\frac{1}{n}\sum_{i=1}^n w_{i}(y_{i}-\mu)>x\Big)\leq \alpha\Big\}.
\end{equation}
Since the probability operator $\mathbb P_{\bw}$ is conditionally on $\by_n$, all the randomness of $q_{\alpha}( \by_{n}-\mu)$ come from $\by_n$. By the symmetric assumption of the reward, the distribution of $y_i-\mu$ is \emph{exactly the same} as the distribution of $w_i(y_i-\mu)$ for Rademacher r.v. $\{w_i\}$.
Then we have
\begin{eqnarray}\label{eqn:1}
&&\mathbb P\Big(\bar{y}_n -\mu > q_{\alpha}( \by_{n}-\mu)\Big)\nonumber\\
&=&\mathbb E_{\bw}\Big[\mathbb P_{\by}\Big(\frac{1}{n}\sum_{i=1}^nw_i(y_i-\mu)>q_{\alpha}((\by_n -\mu) \circ \bw_n))\Big)\Big],
\end{eqnarray}
where $\circ$ is the Hadamard product. By Fubini's theorem, we can interchange the probability operator and expectation operator as follows
\begin{eqnarray}\label{eqn:2}
    &&\mathbb E_{\bw}\Big[\mathbb P_{\by}\Big(\frac{1}{n}\sum_{i=1}^nw_i(y_i-\mu)>q_{\alpha}((\by_n -\mu) \circ \bw_n)\Big)\Big]\nonumber\\
    &=& \mathbb E_{\by}\Big[\mathbb P_{\bw}\Big(\frac{1}{n}\sum_{i=1}^nw_i(y_i-\mu)>q_{\alpha}(\by_n -\mu)\Big)\Big] \leq \alpha,
\end{eqnarray}
where the first inequality is due to the fact that for any arbitrary sign reversal, $q_{\alpha}((\by_n -\mu) \circ \bw_n) = q_{\alpha}(\by_n -\mu) $ based on the definition of $q_{\alpha}$ and the last inequality is from the definition of quantitle. Combining \eqref{eqn:1} and \eqref{eqn:2} together, we conclude that
\begin{equation}\label{ineqn:quantitle}
    \mathbb P\Big(\bar{y}_n -\mu > q_{\alpha}( \by_{n}-\mu)\Big) \leq \alpha.
\end{equation}

\paragraph{Step Two.} We define a good event 
\begin{equation}\label{eqn:e_c}
    \cE = \Big\{\by_n|q_{\alpha}(\by_n -\mu)\leq q_{\alpha(1-\delta)}(\by_n-\bar{y}_n) + \sqrt{\frac{2\log(2/\alpha\delta)}{n}}\varphi(\by_n)\Big\}.
\end{equation}

Together with \eqref{ineqn:quantitle} and union event trick,
\begin{eqnarray*}
&&\mathbb P \Big(\bar{y}_n -\mu > q_{\alpha(1-\delta)}(\by_n-\bar{y}_n) + \sqrt{\frac{2\log(2/\alpha\delta)}{n}}\varphi(\by_n)\Big)\\
&=&\mathbb P \Big(\Big\{\bar{y}_n -\mu > q_{\alpha(1-\delta)}(\by_n-\bar{y}_n) + \sqrt{\frac{2\log(2/\alpha\delta)}{n}}\varphi(\by_n)\Big\} \cap \big(\{\by_n\in\cE\}\cup \{\by_n\in\cE^c\}\big)\Big)\\
&=& \mathbb P \Big(\Big\{\bar{y}_n -\mu > q_{\alpha(1-\delta)}(\by_n-\bar{y}_n) + \sqrt{\frac{2\log(2/\alpha\delta)}{n}}\varphi(\by_n)\Big\} \cap \{\by_n\in\cE\}\Big) \\
&&+ \mathbb P \Big(\Big\{\bar{y}_n -\mu > q_{\alpha(1-\delta)}(\by_n-\bar{y}_n) + (2\log(2/\alpha\delta)/n)^{1/2}\varphi(\by_n)\Big\} \cap \{\by_n\in\cE^c\}\Big)\\
&\leq& \mathbb P \Big(\bar{y}_n -\mu > q_{\alpha}(\by_n-\mu)\Big) + \mathbb P\Big(\by_n\in\cE^c\Big)\\
 &\leq& \alpha + \mathbb P(\by_n\in\cE^c).
\end{eqnarray*}
To bound $\mathbb P(\by_n\in\cE^c)$, we first prove the following claim:
\begin{equation}\label{eqn:subset}
   \text{Claim:} \ \ \ \ \cE^c\subset \Big\{\by_n|\mathbb P_{\bw}\Big(\bar{w}_n(\bar{y}_n-\mu)>\sqrt{\frac{2\log(2/\alpha\delta)}{n}}\varphi(\by_n)\Big)\geq \alpha\delta\Big\},
\end{equation}
where $\bar{w}_n = \sum_{i=1}^n w_i/n$. To show this, we have by the definition of $ q_{\alpha}( \by_{n}-\mu)$ in \eqref{def:quantitle_mean},
\begin{equation*}
    \mathbb P_{\bw}\Big(\frac{1}{n}\sum_{i=1}^n w_{i}(y_{i}-\mu)>q_{\alpha}(\by_n-\mu)\Big) = \alpha.
\end{equation*}
By some simple algebras, we have 
\begin{eqnarray}\label{eqn:w_decompo}
    \frac{1}{n}\sum_{i=1}^n w_{i}(y_{i}-\mu) = \frac{1}{n}\sum_{i=1}^n w_{i}(y_{i}-\bar{y}_n+\bar{y}_n - \mu) = \frac{1}{n}\sum_{i=1}^n w_{i}(y_{i}-\bar{y}_n) + \bar{w}_n(\bar{y}_n-\mu).
\end{eqnarray}
For any $\by_n\in\cE^c$,
\begin{eqnarray*}
     \alpha&=&\mathbb P_{\bw}\Big(\frac{1}{n}\sum_{i=1}^n w_{i}(y_{i}-\mu)>q_{\alpha}(\by_n-\mu)\Big)\\
     &\leq & \mathbb P_{\bw}\Big(\frac{1}{n}\sum_{i=1}^n w_{i}(y_{i}-\mu)>q_{\alpha(1-\delta)}(\by_n-\bar{y}_n) +\sqrt{\frac{2\log (2/\alpha\delta)}{n}} \varphi(\by_n)\Big) \ (\text{by the definition of } \cE^c)\\
     &=& \mathbb P_{\bw}\Big(\frac{1}{n}\sum_{i=1}^n w_{i}(y_{i}-\bar{y}_n) + \bar{w}_n(\bar{y}_n-\mu)>q_{\alpha(1-\delta)}(\by_n-\bar{y}_n) + \sqrt{\frac{2\log(2/\alpha\delta)}{n}}\varphi(\by_n)\Big) \ (\text{by} \ \eqref{eqn:w_decompo})\\
     &\leq&  \mathbb P_{\bw}\Big(\frac{1}{n}\sum_{i=1}^n w_{i}(y_{i}-\bar{y}_n) >q_{\alpha(1-\delta)}(\by_n-\bar{y}_n)\Big) +  \mathbb P_{\bw}\Big(\bar{w}_n(\bar{y}_n-\mu)>\sqrt{\frac{2\log(2/\alpha\delta)}{n}}\varphi(\by_n)\Big)\\
     &\leq& \alpha(1-\delta) + \mathbb P_{\bw}\Big(\bar{w}_n(\bar{y}_n-\mu)>\sqrt{\frac{2\log(2/\alpha\delta)}{n}}\varphi(\by_n)\Big).
\end{eqnarray*}
This proves the claim of \eqref{eqn:subset}.

\paragraph{Step Three.} We start to bound the second term above as follows,
\begin{eqnarray}\label{eqn:p1}
    &&\mathbb P_{\bw}\Big(\bar{w}_n(\bar{y}_n-\mu)>\sqrt{\frac{2\log(2/\alpha\delta)}{n}}\varphi(\by_n)\Big)\\
    &\leq& \mathbb P_{\bw}\Big(|\bar{w}_n(\bar{y}_n-\mu)|>\sqrt{\frac{2\log(2/\alpha\delta)}{n}}\varphi(\by_n)\Big)\nonumber\\
    &\leq& \mathbb P_{\bw} \Big(n|\bar{w}_n|>\sqrt{2n\log(2/\alpha\delta)}\frac{\varphi(\by_n)}{|\bar{y}_n-\mu|}\Big),
\end{eqnarray}   
where the last inequality is actually conditional on the event $\{\bar{y}_n\neq \mu\}$ that holds with probability one. Note that $(w_i+1/2)\sim\text{Bernoulli}(1/2)$ and thus $\sum_{i=1}^n (w_i+1)/2\sim \text{Binomial}(n,1/2)$. Denote $X_n$ is a Binomial($n, 1/2$) random variable. Applying the sharp large deviation bound in Lemma \ref{lemma:large_deviation} with $p_i = 1/2$, we have 
\begin{eqnarray}\label{eqn:p2}
    \mathbb P_{X_n}\Big(X_n-\frac{n}{2}>\sqrt{2n\log(2/\alpha\delta)}\frac{\varphi(\by_n)}{|\bar{y}_n-\mu|}\Big) &\leq& 2\exp\Big(-2\frac{\varphi(\by_n)^2}{(\bar{y}_n-\mu)^2}2n\log(2/\alpha\delta)\frac{1}{n}\Big)\nonumber\\
    &=& 2\exp\Big(-\frac{4\log(2/\alpha\delta)\varphi(y_n)^2}{(\bar{\by}_n-\mu)^2}\Big).
\end{eqnarray}
Putting \eqref{eqn:p1} and \eqref{eqn:p2} together, 
\begin{equation*}
     \mathbb P_{\bw}\Big(\bar{w}_n(\bar{y}_n-\mu)>\sqrt{\frac{2\log(2/\alpha\delta)}{n}}\varphi(\by_n)\Big) \leq 2\exp\Big(-\frac{\log(2/\alpha\delta)\varphi(\by_n)^2}{(\bar{y}_n-\mu)^2}\Big).
\end{equation*}

From \eqref{eqn:subset}, it remains to bound
\begin{eqnarray*}
    \mathbb P\Big(\by_n\in\cE^c\Big) &\leq& \mathbb P_{\by}\Big(\mathbb P_{\bw}\Big(\bar{w}_n(\bar{y}_n-\mu)>\sqrt{\frac{2\log(2/\alpha\delta)}{n}}\varphi(\by_n)\Big)\geq \alpha\delta\Big)\\
    &\leq & \mathbb P_{\by}\Big(2\exp\Big(-\frac{4\log(2/\alpha\delta)\varphi(\by_n)^2}{(\bar{y}_n-\mu)^2}\Big) \geq \alpha\delta\Big)\\
    &=& \mathbb P_{\by}\Big(|\bar{y}_n-\mu| \geq 2\varphi(\by_n)\Big).
\end{eqnarray*}
This reaches 
\begin{eqnarray}
    \mathbb P \Big(\bar{y}_n -\mu > q_{\alpha(1-\delta)}(\by_n-\bar{y}_n) + \sqrt{\frac{2\log(2/\alpha\delta)}{n}}\varphi(\by_n)\Big) \leq \alpha + \mathbb P_{\by_n}\Big(|\bar{y}_n-\mu| \geq \varphi(\by_n)\Big).
\end{eqnarray}
Letting $\varphi(\by_n)$ be a non-negative function such that 
\begin{equation*}
    \mathbb P_{\by}\Big(|\bar{y}_n-\mu |\geq \varphi(\by_n)\Big) \leq \alpha,
\end{equation*}
we have
\begin{equation*}
    \mathbb P \Big(\bar{y}_n -\mu > q_{\alpha(1-\delta)}(\by_n-\bar{y}_n) + \sqrt{\frac{2\log(2/\alpha\delta)}{n}}\varphi(\by_n)\Big) \leq 2\alpha.
\end{equation*}
Redefine $\varphi(\by_n)=2\varphi(\by_n)$ with a little bit abuse of notations. This ends our proof. \hfill $\blacksquare$\\

%%%%%%%%%%%%%%%%%%%%%%%%%%%%%%%%%%%%%%%%%%%%%%%%%%%%%%%%%
\subsection{Proof of Theorem \ref{thm:orlicz_concentra}}\label{proof:concentration}
%%%%%%%%%%%%%%%%%%%%%%%%%%%%%%%%%%%%%%%%%%%%%%%%%%%%%%%%

We start by an upper bound for the $p$-th moment of sum of sub-Weibull random variables with bounded $\psi_{\beta}$-norm. The proof of Lemma \ref{lemma:psi_moment_bound} is deferred to Section \ref{proof:main_lemmas}.
\begin{lemma}\label{lemma:psi_moment_bound}
	Suppose $\{y_i\}_{i=1}^n$ are $n$ independent sub-Weibull random variables satisfying $\|y_i\|_{\psi_{\beta}}\leq \sigma$ with $\beta> 0$. Then for all $\ba=(a_1,\ldots, a_n)\in\mathbb R^n$ and $p\geq 2$, we have
	\begin{eqnarray}\label{eqn:psi_moment_bound}
\Big(\mathbb E\Big|\sum_{i=1}^na_i y_i-\mathbb E(\sum_{i=1}^na_iy_i)\Big|^p\Big)^{\tfrac{1}{p}}
	\leq
	\left\{\begin{array}{ll}
	C_{\beta}\sigma\big(\sqrt{p}\|\ba\|_2+p^{1/\beta}\|\ba\|_{\infty}\big), & \text{ if } 0<\beta<1;\\
	C_{\beta}\sigma\big(\sqrt{p}\|\ba\|_2+p^{1/\beta}\|\ba\|_{\beta^*}\big), & \text{ if } \beta\geq 1.\\	
	\end{array}\right.
	\end{eqnarray}
	where $1/\beta^*+1/\beta = 1$, $C_{\beta}$ are some absolute constants only depending on $\beta$.
\end{lemma}

\begin{remark}
If $0<\beta<1$, \eqref{eqn:psi_moment_bound} is a combination of Theorem 6.2 in \cite{HMO1997} and the fact that the $p$-th moment of a Weibull variable with parameter $\beta$ is of order $p^{1/\beta}$. If $\beta\geq 1$, \eqref{eqn:psi_moment_bound} follows from a combination of Corollaries 2.9 and 2.10 in \cite{talagrand1994supremum}. Continuing with standard symmetrization arguments, we reach the conclusion for general random variables. When $\beta=1$ or 2, \eqref{eqn:psi_moment_bound} coincides with standard moment bounds for a sum of sub-Gaussian and sub-exponential random variables in \cite{RV12}. 
\end{remark}

After we get the $p$-th moment bound in Lemma \ref{lemma:psi_moment_bound}, we can use Markov's inequality to transfer it to a high-probability as follows. For any $t>0$, by Markov's inequality,
\begin{equation*}
\begin{split}
&\mathbb P\Big(\Big|\sum_{i=1}^n a_iy_i-\mathbb E\Big(\sum_{i=1}^na_iy_i\Big)\Big|\geq t \Big)=\mathbb P\Big(\Big|\sum_{i=1}^n a_iy_i-\mathbb E\Big(\sum_{i=1}^na_iy_i\Big)\Big|^p\geq t^p \Big)\\
\leq& \frac{\mathbb E\Big|\sum_{i=1}^n a_iy_i-\mathbb E\Big(\sum_{i=1}^na_iy_i\Big)\Big|^p}{t^p}\leq \frac{C_{\beta}^p\sigma^p\Big(\sqrt{p}\|\ba\|_2+p^{1/\beta}\|\ba\|_{\infty}\Big)^p}{t^p},
\end{split}
\end{equation*}
where the last inequality is from Lemma \ref{lemma:psi_moment_bound}. By setting $t$ such that 
$$\exp(-p) =C_{\beta}^p\sigma^p(\sqrt{p}\|\ba\|_2+p^{1/\beta}\|\ba\|_{\infty})^p/t^p,$$ 
we have for $p\geq 2$,
\begin{equation*}
    \Big|\sum_{i=1}^n a_iy_i-\mathbb E\Big(\sum_{i=1}^na_iy_i\Big)\Big|\leq eC_{\beta}\sigma\Big(\sqrt{p}\|\ba\|_2+p^{1/\beta}\|\ba\|_{\infty}\Big)
\end{equation*}
holds with probability at least $1-\exp(-p)$. Letting $\alpha= \exp(-p)$, we have that for any $0<\alpha<1/e^2$,
\begin{equation*}
    \Big|\sum_{i=1}^n a_iy_i-\mathbb E\Big(\sum_{i=1}^na_iy_i\Big)\Big|\leq C_{\beta}\sigma\Big(\|\ba\|_2(\log\alpha^{-1})^{1/2}+\|\ba\|_{\infty}(\log \alpha^{-1})^{1/\beta}\Big),
\end{equation*}
holds with probability at least $1-\alpha$. This ends the proof. \hfill $\blacksquare$

\subsection{Proof of Theorem \ref{thm:regret_problem_dependent}}\label{proof:main}
We first prove a problem-dependent bound then a problem-independent bound.
\paragraph{Problem-Dependent Bound.} Recall that at round $t+1$, the UCB index used in our algorithm is 
\begin{equation*}
\text{UCB}_k(t) = \bar{y}_{n_{k,t}} + h_{\alpha}(\by_{n_{k,t}}),
\end{equation*}
where $n_{k,t}$ is the number of pulls until round $t+1$ for arm $k$ and 
\begin{equation*}
    h_{\alpha}(\by_{n_{k,t}}) = q_{\alpha/2}\big(\by_{n_{k,t}}-\bar{y}_{n_{k,t}}\big) + \sqrt{\frac{2\log (4/\alpha)}{n_{k,t}}}\varphi(\by_{n_{k,t}}),
\end{equation*}
where
\begin{equation}\label{eqn:varphi}
    \varphi(\by_{n_{k,t}}) = C_{\beta}\sigma \Big(\sqrt{\frac{\log 1/\alpha}{n_{k,t}}} + \frac{(\log 2/\alpha)^{1/\beta}}{n_{k,t}}\Big).
\end{equation}
From Theorem \ref{thm:orlicz_concentra}, for any fixed $n_{k,t} = s$, we know that 
\begin{eqnarray*}
    \mathbb P\Big(\bar{y}_s - \mu_k\geq \varphi(\by_s)\Big)\leq \alpha.
\end{eqnarray*}
From Theorem \ref{thm:non_asy_bootstrap}, for any fixed $n_{k,t} = s$, we have
\begin{equation}\label{eqn:concentration}
    \mathbb P\Big(\mu_k - \bar{y}_{s} >h_{\alpha}(\by_{s})\Big)\leq 2\alpha, \ k\in[K].
\end{equation}

The basic idea is to bound the expected number of pulls $\mathbb E(n_{k,t})$ for sub-optimal arms. To decouple the randomness from the behavior of the UCB algorithm, we define a good event as follows,
\begin{equation}
    \cE_k = \{\mu_1<\min_{t\in[T]}\text{UCB}_1(t)\}\cap\{\bar{y}_{b_k}+h_{\alpha}(\by_{b_k})<\mu_1\}, \ k\in[K],
\end{equation}
where $b_k\in[T]$ is a constant to be chosen later. 

 First, we want to prove the following claim: if event $\cE_k$ happens, then $n_{k,t}\leq b_k$. To show this, we use a contradiction argument. If $n_{k,t}>b_k$, then arm $k$ was pulled more than $b_k$ times over the first $T$ rounds, and so there must exist a round $t\in[T]$ such that $n_{k,t} =b_k$ and $I_t =k$. This implies 
\begin{eqnarray*}
    \text{UCB}_k(t) &=&  \bar{y}_{n_{k,t}} + h_{\alpha}(\by_{n_{k,t}}) =  \bar{y}_{b_k} + h_{\alpha}(\by_{b_k}).
\end{eqnarray*}
From the definition of $\cE_k$, we have
\begin{equation*}
    \bar{y}_{b_k} + h_{\alpha}(\by_{b_k})< \mu_1<\min_{t'\in[T]}\text{UCB}_1(t')\leq \text{UCB}_1(t).
\end{equation*}
This results in a contradiction. Then we can decompose $\mathbb E[n_{k,t}]$ with respect to the event $\cE_k$,
\begin{equation}\label{eqn:one_step_regret}
    \mathbb E [n_{k,t}] = \mathbb E[I(\cE_k)n_{k,t}] +  \mathbb E[I(\cE_k^c)n_{k,t}] \leq b_k+\mathbb P(\cE_k^c) T.
\end{equation}

Second, we will derive an upper bound for $\mathbb P(\cE_k^c) T$. By definition,
\begin{eqnarray}\label{eqn:dec_E_k}
   \mathbb P( \cE_k^c) &=& \mathbb P\Big(\{\mu_1\geq \min_{t\in[T]}\text{UCB}_1(t)\}\cup \{\bar{y}_{b_k}+h_{\alpha}(\by_{b_k})\geq \mu_1\}\Big)\nonumber\\
   &\leq & \underbrace{\mathbb P\Big(\mu_1\geq \min_{t\in[T]}\text{UCB}_1(t)\Big)}_{I_1} + \underbrace{\mathbb P\Big( \bar{y}_{b_k}+h_{\alpha}(\by_{b_k})\geq \mu_1\Big)}_{I_2}.
\end{eqnarray}
To bound $I_1$, we apply the union bound trick as follows,
\begin{eqnarray*}
    \{\mu_1\geq \min_{t\in[T]}\text{UCB}_1(t)\}&\subset& \{\mu_1\geq \min_{s\in[T]}\bar{y}_{s}+ h_{\alpha}(\by_{s})\}\\
    &=& \cup_{s\in[T]}\{\mu_1\geq \bar{y}_s+h_{\alpha}(\by_s)\}.
\end{eqnarray*}
By \ref{eqn:concentration}, it implies
\begin{equation}\label{eqn:dec_E_k1}
    \mathbb P\Big(\mu_1\geq \min_{t\in[T]}\text{UCB}_1(t)\Big)\leq \sum_{s=1}^T\mathbb P\Big(\mu_1\geq \bar{y}_s + h_{\alpha}(\by_s)\Big) \leq 2\alpha T.
\end{equation}

To bound $I_2$, the key step is to derive an sharp upper bound for threshold $h_{\alpha}(\by_{b_k})$. Next lemma presents an upper bound for the multiplier bootstrapped quantile which is the main part of $h_{\alpha}(\by_{b_k})$. The proof is deferred to Section \ref{proof:lemma:quantile_bound}.

\begin{lemma}\label{lemma:quantile_bound}
Suppose $\{y_i-\mu\}_{i=1}^n$ follows sub-Weibull distribution with $\|y_i-\mu\|_{\psi_{\beta}}\leq \sigma$ and $\{w_i\}_{i=1}^n$ are i.i.d Rademacher random variables independent of $y_i$.
Then we have
\begin{eqnarray}\label{eqn:quantitle_ineq}
    \mathbb P \Big(\frac{1}{n}\sum_{i=1}^n (w_i-\bar{w})(y_i-\mu)\leq C_{\beta}\sigma\Big(\sqrt{\frac{\log (1/\alpha)}{n}}+\frac{(\log( 1/\alpha))^{1/\beta}}{n}\Big)\Big)\geq 1-\alpha.
\end{eqnarray}
\end{lemma}

By the definition of $q_{\alpha/2}(\by_{b_k}-\bar{y}_{b_k})$ in \eqref{def:quantile}, we have 
\begin{eqnarray}
    q_{\alpha/2}(\by_{b_k}-\bar{y}_{b_k}) \leq C_{\beta}\sigma\Big(\sqrt{\frac{\log (2/\alpha)}{b_k}}+\frac{(\log (2/\alpha))^{1/\beta}}{b_k}\Big),
\end{eqnarray}
with probability at least $1-\alpha/2$. Recall that 
\begin{eqnarray}
    \sqrt{\frac{2\log(4/\alpha)}{b_k}}\varphi(\by_{b_k}) = \sqrt{\frac{2\log (4/\alpha)}{b_k}}\Big(\sqrt{\frac{\log (1/\alpha)}{b_k}}+\frac{(\log (1/\alpha))^{1/\beta}}{b_k}\Big).
\end{eqnarray}
Overall, we have 
\begin{eqnarray}\label{eqn:h_bound}
    h_{\alpha}(\by_{b_k}) &=&  q_{\alpha/2}(\by_{b_k}-\bar{y}_{b_k}) +  \sqrt{\frac{2\log(4/\alpha)}{b_k}}\varphi(\by_{b_k}) \\
    &\leq& 2C_{\beta}\sigma\Big(\sqrt{\frac{\log (2/\alpha)}{b_k}}+\frac{(\log( 2/\alpha))^{1/\beta}}{b_k}\Big),
\end{eqnarray}
with probability $1-\alpha/2$ as long as $b_k\geq 2\log(4/\alpha)/(C_{\beta}^2\sigma^2)$. 

For two events $\cA$ and $\cB$, we have 
\begin{equation}\label{eqn:union_event}
    \mathbb P(\cA) = \mathbb P(\cA\cap\cB^c) + \mathbb P(\cA\cap\cB)\leq  \mathbb P(\cA\cap\cB) +\mathbb P(\cB^c).
\end{equation}
Next we define an event $\cB_k = \{  h_{\alpha}(\by_{b_k})\leq \Delta_k/2\}$, where $\Delta_k = \mu_1 - \mu_k$. We decompose $I_2$ with respect to $\cB_k$ following the union event rule \eqref{eqn:union_event},
\begin{eqnarray*}
    &&\mathbb P\Big(\bar{y}_{b_k} + h_{\alpha}(\by_{b_k})\geq \mu_1\Big)\\
    &=& \mathbb P\Big(\bar{y}_{b_k} + h_{\alpha}(\by_{b_k})-\mu_k\geq \mu_1-\mu_k\Big)\\
    &\leq&  \mathbb P\Big(\bar{y}_{b_k} -\mu_k\geq\Delta_k - h_{\alpha}(\by_{b_k})\cap\cB_k\Big) + \mathbb P(\cB_k^c)\\
    &\leq & \mathbb P\Big(\bar{y}_{b_k}-\mu_k\geq \frac{\Delta_k}{2}\cap\cB_k\Big) + \mathbb P(\cB_k^c)\\
    &\leq & \mathbb P\Big(\bar{y}_{b_k}-\mu_k\geq \frac{\Delta_k}{2}\Big) + \mathbb P(\cB_k^c).
\end{eqnarray*}

To bound the first part, we reuse the concentration inequality in Theorem \ref{thm:orlicz_concentra} such that,
\begin{eqnarray}\label{bound_first_part}
    \mathbb P \Big(\bar{y}_{b_k}-\mu_k\geq \frac{\Delta}{2}\Big)\leq \exp\Big(-\min\Big[\Big(\frac{\Delta_k}{C_{\beta}\sigma}\Big)^2b_k, \Big(\frac{\Delta_kb_k}{4C_{\beta}\sigma}\Big)^{\beta}\Big]\Big).
\end{eqnarray}

To bound the second part, we bound $ \mathbb P(\cB_k^c)$ in three steps,
\begin{enumerate}
    \item By \eqref{eqn:h_bound}, we have 
    \begin{eqnarray}\label{eqn:B_k}
        \mathbb P(\cB_k^c) &=& \mathbb P\Big( h_{\alpha}(\by_{b_k})> \Delta_k/2\Big)\nonumber\\
        &\leq& \mathbb P\Big( 2C_{\beta}\sigma\Big(\sqrt{\frac{\log (2/\alpha)}{b_k}}+\frac{(\log (2/\alpha))^{1/\beta}}{b_k}\Big)> \Delta_k/2\Big) + \alpha/2.
    \end{eqnarray}
    \item To ensure that $2C_{\beta}\sigma\sqrt{\log (2/\alpha)/b_k} \leq \Delta_k/4$, we need 
    \begin{eqnarray*}
        b_k\geq \Big(\frac{8C_\beta\sigma}{\Delta_k}\Big)^2\log (2/\alpha).
    \end{eqnarray*}
 To ensure that $2C_{\beta}\sigma(\log (2/\alpha))^{(1/\beta)}/b_k\leq \Delta_k/4$, we need 
    \begin{eqnarray*}
        b_k\geq \frac{8C_{\beta}\sigma(\log (2/\alpha))^{(1/\beta)}}{\Delta_k}.
    \end{eqnarray*}
    \item Then if we choose $b_k$ as 
    \begin{eqnarray}\label{eqn:b_k}
        b_k =  \Big(\frac{8C_\beta\sigma}{\Delta_k}\Big)^2\log (2/\alpha) +  \frac{8C_{\beta}\sigma(\log (2/\alpha))^{1/\beta}}{\Delta_k},
    \end{eqnarray}
    we have 
    \begin{eqnarray}\label{eqn:B_k2}
         \mathbb P\Big( 2C_{\beta}\sigma\Big(\sqrt{\frac{\log (2/\alpha)}{b_k}}+\frac{(\log (2/\alpha))^{1/\beta}}{b_k}\Big)> \Delta_k/2\Big) = 0.
    \end{eqnarray}
\end{enumerate}
Combining \eqref{eqn:B_k} and \eqref{eqn:B_k2},  we conclude that when $b_k$ is choose as in \eqref{eqn:b_k}, we have 
\begin{eqnarray}\label{bound_second_part}
     \mathbb P(\cB_k^c)  \leq \alpha/2.
\end{eqnarray}

Combing \eqref{bound_first_part} and \eqref{bound_second_part}, we have 
\begin{eqnarray}\label{eqn:dec_E_k2}
    \mathbb P\Big(\bar{y}_{b_k} + h_{\alpha}(\by_{b_k})\geq \mu_1\Big) \leq \exp\Big(-\min\Big[\Big(\frac{\Delta_k}{C_{\beta}\sigma}\Big)^2b_k, \Big(\frac{\Delta_kb_k}{4C_{\beta}\sigma}\Big)^{\beta}\Big]\Big) + \alpha/2,
\end{eqnarray}
when $b_k$ is chosen as below
\begin{eqnarray*}
      b_k =  \Big(\frac{8C_\beta\sigma}{\Delta_k}\Big)^2\log (1/\alpha) +  \frac{8C_{\beta}\sigma(\log (2/\alpha))^{1/\beta}}{\Delta_k}.
\end{eqnarray*}

Combining \eqref{eqn:dec_E_k}, \eqref{eqn:dec_E_k1} and \eqref{eqn:dec_E_k2} together,
\begin{eqnarray}\label{eqn:bound_E_k}
    \mathbb P(\cE_k^c) &\leq& 2T\alpha + \exp\Big(-\min\Big[\Big(\frac{\Delta_k}{C_{\beta}\sigma}\Big)^2b_k, \Big(\frac{\Delta_kb_k}{4C_{\beta}\sigma}\Big)^{\beta}\Big]\Big) + \alpha/2\nonumber\\
    &\leq& 2T\alpha + \exp\Big(-\min\Big[\Big(\frac{\Delta_k}{C_{\beta}\sigma}\Big)^2\Big(\frac{8C_\beta\sigma}{\Delta_k}\Big)^2\log (2/\alpha), \Big(\frac{\Delta_k}{4C_{\beta}\sigma} \frac{8C_{\beta}\sigma(\log (2/\alpha))^{1/\beta}}{\Delta_k}\Big)^{\beta}\Big]\Big) + \alpha/2\nonumber\\
    &=&2T\alpha + \exp\Big(-\min(64, 2^{\beta})\log (2/\alpha)\Big) + \alpha/2.
\end{eqnarray}
Plugging \eqref{eqn:b_k}, \eqref{eqn:bound_E_k} into \eqref{eqn:one_step_regret},
\begin{eqnarray*}
    \mathbb E[n_{k,t}] &\leq& b_k + \mathbb P(\cE_k^c)T\\
    &=&  \Big(\frac{8C_\beta\sigma}{\Delta_k}\Big)^2\log (2/\alpha) +  \frac{8C_{\beta}\sigma(\log (2/\alpha))^{1/\beta}}{\Delta_k} + 2T^2\alpha + T\alpha^{\min(64, 2^{\beta})} + T\alpha/2.
\end{eqnarray*}
By choosing $\alpha = 2/T^2$, we have 
\begin{eqnarray}\label{bound:N}
     \mathbb E[n_{k,t}] \leq  \Big(\frac{8C_{\beta}\sigma}{\Delta_k}\Big)^2 2\log T + \frac{8C_{\beta}\sigma}{\Delta_k}(2\log T)^{1/\beta}+4,
\end{eqnarray}
since $1-2\min(64, 2^{\beta})<0$ for $\beta>0$. Finally, the cumulative regret is upper bounded by 
\begin{eqnarray}
    R(T) &=& \sum_{k=2}^K \Delta_k \mathbb E[n_{k,t}]\\
    &\leq& \sum_{k=2}^K 128(C_{\beta}\sigma)^2\frac{\log T}{\Delta_k} + 8C_{\beta}\sigma K(2\log T)^{1/\beta} + 4\sum_{k=2}^K\Delta_k.
\end{eqnarray}
 This ends the proof. 
\paragraph{Problem-Independent Bound.} First we let $\Delta>0$ as a threshold which will be specified later. Then we decompose $R(T)$ with respect to the value of $\Delta$ as follows,
\begin{eqnarray}\label{eqn:R_T}
    R(T) &=& \sum_{k=2}^K \Delta_k \mathbb E[n_{k,t}]\nonumber\\
    &=&  \sum_{k:\Delta_k<\Delta}\Delta_k \mathbb E[n_{k,t}] + \sum_{k:\Delta_k\geq\Delta}\Delta_k \mathbb E[n_{k,t}]\nonumber \\
    &\leq &  T\Delta +\sum_{k:\Delta_k\geq \Delta}\Big(128(C_{\beta}\sigma)^2\frac{\log T}{\Delta_k} + 8C_{\beta}\sigma (2\log T)^{1/\beta} + 4\Delta_k\Big)\nonumber\\
    &\leq& 8C_{\beta}\sigma K(2\log T)^{1/\beta} +  4\sum_{k=2}^K\Delta_k+128(C_{\beta}\sigma)^2\frac{K\log T}{\Delta} + T\Delta,
\end{eqnarray}
where the first inequality is from \eqref{bound:N}. Letting $128(C_{\beta}\sigma)^2\frac{K\log T}{\Delta} = T\Delta$, we have \begin{eqnarray}\label{eqn:delta}
    \Delta = (128C_{\beta}^2\sigma^2\frac{K\log T}{T})^{1/2}.
\end{eqnarray}
Plugging \eqref{eqn:delta} back into \eqref{eqn:R_T}, we have
\begin{eqnarray*}
    R(T)\leq 2*128^{1/2}C_{\beta}\sigma\sqrt{TK\log T} +4\sum_{k=1}^K \Delta_k + 8C_{\beta}\sigma K(2\log T)^{1/\beta}.
\end{eqnarray*}
When $T\geq 2^{2/\beta-3}K(\log T)^{2/ \beta - 1}$, we have
\begin{eqnarray*}
    R(T)&\leq& 32\sqrt{2}C_{\beta}\sigma\sqrt{TK\log T} +4\sum_{k=2}^K \Delta_k\\
    &\leq& 32\sqrt{2}C_{\beta}\sigma\sqrt{TK\log T} +4K\mu_1^*.
\end{eqnarray*}
This ends the proof.   \hfill $\blacksquare$\\

%%%%%%%%%%%%%%%%%%%%%%%%%%%%%%%%%%
\section{Proofs of Main Lemmas}\label{proof:main_lemmas}
%%%%%%%%%%%%%%%%%%%%%%%%%%%%%%%%%%
In this section, we provide the proofs of Lemmas \ref{lemma:psi_moment_bound} and \ref{lemma:quantile_bound}.

\subsection{Proof of Lemma \ref{lemma:psi_moment_bound}}

Without loss of generality, we assume $\|x_i\|_{\psi_{\beta}}=1$ and $\mathbb{E}x_i = 0$ throughout this proof. Let $\beta = (\log 4)^{1/\beta}$. For notation simplicity, we define $ \|x\|_p = (\mathbb E|x|^p)^{1/p}$ for a random variable $X$. The following step is to estimate the moment of linear combinations of variables $\{x_i\}_{i=1}^n$.

According to the symmetrization inequality (e.g., Proposition 6.3 of \cite{ledoux2013probability}), we have 
\begin{equation}\label{ineqn:contration}
    \Big\|\sum_{i=1}^n a_ix_i\Big\|_p\leq 2\Big\|\sum_{i=1}^n a_i\varepsilon_i x_i\Big\|_p = 2\Big\|\sum_{i=1}^na_i\varepsilon_i|x_i|\Big\|_p,
\end{equation}
where $\{\varepsilon_i\}_{i=1}^n$ are independent Rademacher random variables and we notice that $\varepsilon_i x_i$ and $\varepsilon_i|x_i|$ are identically distributed. By triangle inequality, 
\begin{eqnarray}\label{ineqn:contration_2}
2\Big\|\sum_{i=1}^na_i\varepsilon_i|x_i|\Big\|_p&\leq& 2\Big\|\sum_{i=1}^na_i\varepsilon_i|x_i-\beta+\beta|\Big\|_p\nonumber\\
&\leq& 2\Big\|\sum_{i=1}^na_i\varepsilon_i|x_i-\beta|\Big\|_p + 2\Big\|\sum_{i=1}^na_i\varepsilon_i\beta\Big\|_p.
\end{eqnarray}

Next, we will bound the second term of the RHS of \eqref{ineqn:contration_2}. In particular, we will utilize Khinchin-Kahane inequality, whose formal statement is included in Lemma \ref{lemma:Khi} for the sake of completeness. From Lemma \ref{lemma:Khi} we have
\begin{eqnarray}\label{ineqn:contration_3}
    \Big\|\sum_{i=1}^n a_i\varepsilon_i\beta\Big\|_p &\leq& \Big(\frac{p-1}{2-1}\Big)^{1/2} \Big\|\sum_{i=1}^n a_i\varepsilon_i\beta\Big\|_2\nonumber\\
    &\leq& \beta\sqrt{p}\Big\|\sum_{i=1}^n a_i\varepsilon_i\Big\|_2.
\end{eqnarray}
Since $\{\varepsilon_i\}_{i=1}^n$ are independent Rademacher random variables, some simple calculations implies
\begin{eqnarray}\label{ineqn:contration_4}
    \Big(\mathbb E\Big(\sum_{i=1}^n \varepsilon_i a_i\Big)^2\Big)^{1/2}&=& \Big(\mathbb E\Big(\sum_{i=1}^n \varepsilon_i^2 a_i^2+2\sum_{1\leq i<j\leq n}\varepsilon_i\varepsilon_j a_i a_j\Big)\Big)^{1/2}\nonumber\\
    &=& \Big(\sum_{i=1}^n a_i^2\mathbb E\varepsilon_i^2 + 2\sum_{1\leq i<j\leq n}a_ia_j\mathbb E\varepsilon_i\mathbb E\varepsilon_j\Big)^{1/2}\nonumber\\
    &=& \Big(\sum_{i=1}^n a_i^2\Big)^{1/2} = \|\ba\|_2.
\end{eqnarray}
Combining inequalities \eqref{ineqn:contration_2}-\eqref{ineqn:contration_4},
\begin{eqnarray}\label{eqn:proof_1}
2\Big\|\sum_{i=1}^na_i\varepsilon_i|x_i|\Big\|_p \leq 2\Big\|\sum_{i=1}^n a_i\varepsilon_i|x_i-\beta|\Big\|_p + 2\beta \sqrt{p}\|\ba\|_2.
\end{eqnarray}
Let $\{y_i\}_{i=1}^n$ be independent symmetric random variables satisfying $\mathbb P(|y_i|\geq t)=\exp(-t^{\beta})$ for all $t\geq 0$. Then we have 
\begin{eqnarray*}
\mathbb P(|x_i-\beta|\geq t) &\leq& \mathbb P(x_i\geq t+\beta) + \mathbb P(x_i\leq \beta-t) \\
&\leq& 2\mathbb P\left(\exp(|x_i|^\beta)\geq \exp((t+\beta)^\beta)\right) \\
&\leq& 2(\mathbb{E} |x_i|^{\beta}) \cdot \exp(-(t+\beta)^\beta) \\
&\leq& 4\exp(-(t+\beta)^{\beta})\\
&\leq& 4\exp(-t^{\beta}-\beta^{\beta}) = \mathbb P(|y_i|\geq t),
\end{eqnarray*}
which implies
\begin{equation}\label{eqn:proof_2}
    \Big\|\sum_{i=1}^n a_i\varepsilon_i|x_i-\beta|\Big\|_p\leq \Big\|\sum_{i=1}^n a_i \varepsilon_i y_i\Big\|_p = \Big\|\sum_{i=1}^n a_i  y_i\Big\|_p,
\end{equation}
since $\varepsilon_i y_i$ and $y_i$ have the same distribution due to symmetry. 
Combining \eqref{eqn:proof_1} and \eqref{eqn:proof_2} together, we reach
\begin{equation}\label{eqn:proof_B1}
    \Big\|\sum_{i=1}^n a_ix_i\Big\|_p\leq 2\beta \sqrt{p}\|\ba\|_2 + 2\Big\|\sum_{i=1}^n a_iy_i\Big\|_p.
\end{equation}

For $0<\beta <1$, it follows Lemma \ref{lemma:alpha2} that
\begin{equation}\label{eqn:proof_B2}
    \Big\|\sum_{i=1}^n a_i y_i\Big\|_p \leq C_{\beta}(\sqrt{p}\|\ba\|_2 + p^{1/\beta}\|\ba\|_{\infty}),
\end{equation}
where $C_{\beta}$ is some absolute constant only depending on $\beta$.

For $\beta\geq 1$, we will combine Lemma \ref{lemma:alpha_1} and the method of the integration by parts to pass from tail bound result to moment bound result. Recall that for every non-negative random variable $x$, integration by parts yields the identity 
\begin{equation*}
    \mathbb E x = \int_{0}^{\infty}\mathbb P(x\geq t)dt.
\end{equation*}
Applying this to $x = |\sum_{i=1}^n a_i y_i|^p$ and changing the variable $t = t^p$, then we have 
\begin{eqnarray}\label{ineqn:integral}
    \mathbb E|\sum_{i=1}^n a_iy_i|^p &=& \int_{0}^{\infty} \mathbb P\Big(|\sum_{i=1}^n a_iy_i|\geq t\Big) p t^{p-1}dt\nonumber\\
    &\leq &\int_0^{\infty} 2\exp\Big(-c\min\Big(\frac{t^2}{\|\ba\|_2^2}, \frac{t^{\beta}}{\|\ba\|_{\beta^*}^{\beta}}\Big)\Big)pt^{p-1}dt,
\end{eqnarray}
where the inequality is from Lemma \ref{lemma:alpha_1} for all $p\geq 2$ and $1/\beta+1/\beta^* = 1$.
In this following, we bound the integral in three steps:
\begin{enumerate}
    \item If $\frac{t^2}{\|\ba\|_2^2}\leq \frac{t^{\beta}}{\|\ba\|_{\beta^*}^{\beta}}$, \eqref{ineqn:integral} reduces to
    \begin{equation*}
          \mathbb E|\sum_{i=1}^n a_iy_i|^p\leq 2p\int_{0}^{\infty}\exp\Big(-c\frac{t^2}{\|\ba\|_2^2}\Big)\Big)t^{p-1}dt.
    \end{equation*}
    Letting $t' = ct^2/\|\ba\|_2^2$, we have 
    \begin{eqnarray*}
        2p\int_{0}^{\infty}\exp\Big(-c\frac{t^2}{\|\ba\|_2^2}\Big)\Big)t^{p-1}dt& =& \frac{p\|\ba\|_2^p}{c^{p/2}}\int_0^{\infty}e^{-t'}t'^{p/2-1}dt'\\
        &=&\frac{p\|\ba\|_2^p}{c^{p/2}} \Gamma(\frac{p}{2})\leq \frac{p\|\ba\|_2^p}{c^{p/2}} (\frac{p}{2})^{p/2},
    \end{eqnarray*}
    where the second equation is from the density of Gamma random variable. Thus,
    \begin{equation}\label{ineqn:moment1}
        \Big(\mathbb E|\sum_{i=1}^n a_iy_i|^p\Big) ^{\tfrac{1}{p}} \leq \frac{p^{1/p}}{(2c)^{1/2}}\sqrt{p}\|\ba\|_2 \leq \frac{\sqrt{2}}{\sqrt{c}}\sqrt{p}\|\ba\|_2.
    \end{equation}

    \item If $\frac{t^2}{\|\ba\|_2^2}> \frac{t^{\beta}}{\|\ba\|_{\beta^*}^{\beta}}$, \eqref{ineqn:integral} reduces to
    \begin{equation*}
         \mathbb E|\sum_{i=1}^n a_iy_i|^p\leq 2p\int_{0}^{\infty}\exp\Big(-c\frac{t^{\beta}}{\|\ba\|_{\beta^*}^{\beta}}\Big)\Big)t^{p-1}dt.
    \end{equation*}
    Letting $t' = ct^{\beta}/\|\ba\|_{\beta^*}^{\beta}$, we have
    \begin{eqnarray*}
        2p\int_{0}^{\infty}\exp\Big(-c\frac{t^{\beta}}{\|\ba\|_{\beta^*}^{\beta}}\Big)\Big)t^{p-1}dt& =& \frac{2p\|\ba\|_{\beta^*}^p}{\beta c^{p/{\beta}}}\int_0^{\infty}e^{-t'}t'^{p/{\beta}-1}dt'\\
        &=&\frac{2}{\beta}\frac{p\|\ba\|_{\beta^*}^p}{c^{p/{\beta}}} \Gamma(\frac{p}{{\beta}})\leq \frac{2}{\beta}\frac{p\|\ba\|_{\beta^*}^p}{c^{p/{\beta}}} (\frac{p}{{\beta}})^{p/{\beta}}.
    \end{eqnarray*}
    Thus, 
     \begin{equation}\label{ineqn:moment2}
        \Big(\mathbb E|\sum_{i=1}^n a_iy_i|^p\Big) ^{\tfrac{1}{p}} \leq \frac{{2p}^{1/p}}{(c\beta)^{1/\beta}}p^{1/\beta}\|\ba\|_{\beta^*} \leq \frac{4}{(c\beta)^{1/\beta}}p^{1/\beta}\|\ba\|_{\beta^*} .
    \end{equation}
    \item Overall, we have the following by combining \eqref{ineqn:moment1} and \eqref{ineqn:moment2},
    \begin{eqnarray*}
    \Big(\mathbb E|\sum_{i=1}^n a_iy_i|^p\Big) ^{\tfrac{1}{p}}\leq \max\Big(\sqrt{\frac{2}{c}},\frac{4}{(c\beta)^{1/\beta}}\Big)\Big(\sqrt{p}\|\ba\|_2 + p^{1/\beta}\|\ba\|_{\beta^*} \Big).
    \end{eqnarray*}
    After denoting $C_{\beta} = \max\Big(\sqrt{\frac{2}{c}},\frac{4}{(c\beta)^{1/\beta}}\Big)$, we reach
    \begin{equation}\label{eqn:proof_B3}
         \Big\|\sum_{i=1}^n a_i y_i\Big\|_p \leq C_{\beta}\Big(\sqrt{p}\|\ba\|_2 + p^{1/\beta}\|\ba\|_{\beta^*} \Big).
    \end{equation}
\end{enumerate}
Since $0 < \beta < 1$, the conclusion can be reached by combining \eqref{eqn:proof_B1},\eqref{eqn:proof_B2} and \eqref{eqn:proof_B3}. \hfill $\blacksquare$\\

\subsection{Proof of Lemma \ref{lemma:quantile_bound}}\label{proof:lemma:quantile_bound}
 Note that with probability one, 
\begin{eqnarray*}
    &&\sum_{i=1}^n (w_i-\bar{w})^2 = \sum_{i=1}^n w_i^2-n\bar{w} - n(1-\bar{w})\leq n,\\
    &&\max_{i}(w_i-\bar{w})\leq 1.
\end{eqnarray*}
We define a good event $\cE$ as follows
\begin{equation}
    \cE = \big\{\sum_{i=1}^n (w_i-\bar{w})^2\leq n\big\} \cup \big\{\max_{i}(w_i-\bar{w})\leq 1\big\}.
\end{equation}
 Then we decompose \eqref{eqn:quantitle_ineq} conditional on $\cE$,
\begin{eqnarray*}
    &&  \mathbb P \Big(\frac{1}{n}\sum_{i=1}^n (w_i-\bar{w})(y_i-\mu)\geq C_{\beta}\sigma\Big(\sqrt{\frac{\log 1/\alpha}{n}}+\frac{(\log 1/\alpha)^{1/\beta}}{n}\Big)\\
    &=& \mathbb P \Big(\frac{1}{n}\sum_{i=1}^n (w_i-\bar{w})(y_i-\mu)\geq C_{\beta}\sigma\Big(\sqrt{\frac{\log 1/\alpha}{n}}+\frac{(\log 1/\alpha)^{1/\beta}}{n}|\cE\Big)\Big)\mathbb P(\cE)\\
    &&+ \mathbb P \Big(\frac{1}{n}\sum_{i=1}^n (w_i-\bar{w})(y_i-\mu)\geq C_{\beta}\sigma\Big(\sqrt{\frac{\log 1/\alpha}{n}}+\frac{(\log 1/\alpha)^{1/\beta}}{n}|\cE^c\Big)\Big)\mathbb P(\cE^c)\\
    &\leq& \mathbb P \Big(\frac{1}{n}\sum_{i=1}^n (w_i-\bar{w})(y_i-\mu)\geq C_{\beta}\sigma\Big(\sqrt{\frac{\log 1/\alpha}{n}}+\frac{(\log 1/\alpha)^{1/\beta}}{n}|\cE\Big)\Big)\\
    &\leq &   \mathbb P \Big(\frac{1}{n}\sum_{i=1}^n (w_i-\bar{w})(y_i-\mu)\geq C_{\beta}\sigma\Big(\frac{(\log 1/\alpha)^{1/2}}{n}\sqrt{\sum_{i=1}^n(w_i-\bar{w})^2}+\frac{(\log 1/\alpha)^{1/\beta}}{n}\max_{i}(w_i-\bar{w})|\cE\Big)\Big)\\
    &\leq&\alpha,
\end{eqnarray*}
where the first inequality is from $\mathbb P(\cE^c) = 0$, the second inequality is from the independence of $w_i$ and $y_i$, the third inequality is from the concentration inequality in Theorem \ref{thm:orlicz_concentra}. This ends the proof.   \hfill $\blacksquare$\\

\section{Monte Carlo Approximations}\label{sec:monte_carlo_approximation}

Suppose $n_{k, t}$ is the number of rewards associated with arm $k$ until round $t$. Practically, we could use Monte Carlo quantile approximation to calculate the multiplier bootstrapped quantile $q_{\alpha}(\by_{n_{k, t}}-\bar{y}_{n_{k, t}})$. Let $\{\bw_n^{(1)}, \ldots, \bw_n^{(B)}\}$ denote $B$ sets of independent random weight vectors and define
\begin{equation}\label{eqn:monte_carlo}
    \tilde{q}_{\alpha} ( \by_{n}-\bar{y}_{n}, \bw^B):=\inf\Big\{x\in\mathbb R \big|\frac{1}{B} \sum_{b=1}^B \mathbf{I}\{\frac{1}{n}\sum_{i=1}^{n} w_{i}^{(b)}(y_{i}-\bar{y}_{n})\geq x\}\leq \alpha\Big\},
\end{equation}
where $B$ is the number of bootstrap repetitions and $\bw^B = (\bw_n^{(1)}, \ldots, \bw_n^{(B)})$. Then the UCB index for arm $k\in[K]$ can be written as
\begin{equation}\label{eqn:UCB_practical}
    \text{UCB}_k(t) = \bar{y}_{n_{k,t}} +\tilde{q}_{\alpha(1-\delta)}(\by_{n_{k, t}}-\bar{y}_{n_{k, t}}, \bw^B) + \sqrt{\frac{2\log (2/\alpha\delta)}{n_{k, t}}}\varphi(\by_{n_{k, t}}).
\end{equation}
The decision-makers choose to pull arm $I_{t+1} = \argmax_{k\in[K]}\text{UCB}_k(t)$. If $\text{UCB}_k(t) = \text{UCB}_{k'}(t)$ for $k\neq k'$, the tie is broken by a fixed rule that is chosen randomly in advance. Next theorem controls the approximation error of the bootstrapped quantile. 

\begin{theorem}[Monte Carlo Quantile Approximation]
Suppose the same conditions in Theorem \ref{thm:non_asy_bootstrap} hold. We have 
\begin{eqnarray*}
     \mathbb P_{\by, \bw}(\bar{y}_n-\mu>\tilde{q}_{\alpha} ( \by_{n}-\bar{y}_{n}, \bw^B) + \sqrt{\log(2/\alpha\delta)/n}\varphi(\by_n))
     \leq \alpha + \frac{\left \lfloor{B\alpha}\right \rfloor  + 1}{B+1}\leq 2\alpha + \frac{1}{B+1},
\end{eqnarray*}
where $\tilde{q}_{\alpha} ( \by_{n}-\bar{y}_{n}, \bw^B)$ is the Monte Carlo approximated quantile defined in (D.1). 
\end{theorem}
By replacing the true quantile $q_{\alpha}$ by a MC quantile $\tilde{q}_{\alpha}^B$ based on $B$ i.i.d bootstrapped weights, we lose at most $1/(B+1)$ for the confidence level. 

\emph{Proof Sketch.} The proof is similar to the proof of Theorem \ref{thm:non_asy_bootstrap} except for the control of i.i.d approximation error. First, we define 
\begin{equation*}
    \tilde{q}_{\alpha} ( \by_{n}-\mu, \bw^B):=\inf\Big\{x\in\mathbb R \big|\frac{1}{B} \sum_{b=1}^B \mathbf{I}\{\frac{1}{n}\sum_{i=1}^{n} w_{i}^{(b)}(y_{i}-\mu)\geq x\}\leq \alpha\Big\}.
\end{equation*}
By using the similar symmetry properties as we did in \eqref{eqn:1} and \eqref{eqn:2}, we have 
\begin{eqnarray*}
    &&\mathbb E_{\bw^B}\mathbb P_{\by}\Big(\frac{1}{n}\sum_{i=1}^nw_i(y_i-\mu)>  \tilde{q}_{\alpha} ( \by_{n}-\mu, \bw^B)\Big)\\
    &=& \mathbb E_{\bw} \mathbb E_{\bw^B}\mathbb P_{\by}\Big(\frac{1}{n}\sum_{i=1}^nw_i(y_i-\mu)>  \tilde{q}_{\alpha} ( (\by_{n}-\mu)\circ \bw_n, \bw^B)\Big)\\
    &=& \mathbb E_{\by} \mathbb P_{\bw,\bw^B}\Big(\frac{1}{n}\sum_{i=1}^nw_i(y_i-\mu)>  \tilde{q}_{\alpha} ( \by_{n}-\mu, \bw^B \cdot \diag(\bw_n))\Big)\\
    &=& \mathbb E_{\by} \mathbb P_{\bw,\bw^B}\Big(\frac{1}{n}\sum_{i=1}^nw_i(y_i-\mu)>  \tilde{q}_{\alpha} ( \by_{n}-\mu, \bw^B )\Big)\\
     &=& \mathbb E_{\by} \mathbb P_{\bw,\bw^B}\Big(\sum_{b=1}^B \mathbf{I}\{\frac{1}{n}\sum_{i=1}^{n} w_{i}^{(b)}(y_{i}-\mu)\geq x\}\leq \alpha\Big)\leq  \frac{\left \lfloor{B\alpha}\right \rfloor  + 1}{B+1},
\end{eqnarray*}
where the last inequality can be derived from Lemma 1 in \citep{romano2005exact}. The rest of the proof will follow step two in the proof of Section \ref{proof:non_asy}. \hfill $\blacksquare$\\

%%%%%%%%%%%%%%%%%%%%%%%%%%%%%%%%%%%%%%%%%%%%%%
\section{Additional Experimental Results and Implementation Details }\label{sec:addition_simu}
%%%%%%%%%%%%%%%%%%%%%%%%%%%%%%%%%%%%%%%%%%%%%%%%
In Section \ref{sec:multi_detail}, we present the implementation details for multi-armed bandits. In Section \ref{sec:linear_bandit}, we present the implementation details for linear bandits. In Section \ref{subsec:dis}, we present formal definitions for logistic distribution and truncated-normal distribution.

\subsection{Multi-armd Bandit} \label{sec:multi_detail}
For UCB1, at each round, the action is selected as 
$$
\argmax_{k\in[K]}\frac{1}{n_k}\sum_{s=1}^{n_k}y_s^k + \hat{\sigma}\sqrt{\frac{2\log(1/\alpha)}{n_k}}.
$$
For Jeffery-TS, at each round, the parameter is sampled from 
$$
\mathbb N\Big(\frac{1}{n_k}\sum_{s=1}^{n_k}y_s^k, \hat{\sigma}^2/n_k\Big).
$$ 
Here, $\hat{\sigma}$ is the upper bound on the estimator of standard deviation, $\{y_s^k\}$ are the reward associated with arm $k$ and $n_k$ is the number of reward associated with arm $k$. For notation simplicity, we ignore their dependency on round $t$. 

\begin{figure}[h]
	\centering
		\vspace{0.1cm}
		\includegraphics[scale=0.5]{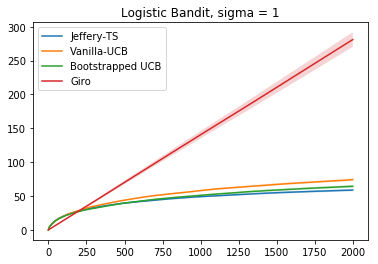}
		\includegraphics[scale=0.5]{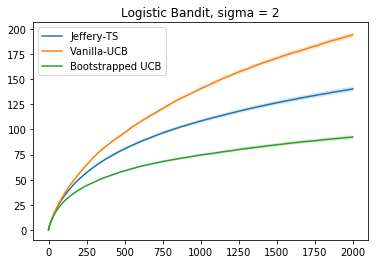}

		\vspace{-0.1in}
\caption{Cumulative regret for logistic bandit. The left panel is for $\hat{\sigma} = 1$, and the right panel is for $\hat{\sigma} = 2$.}\label{plt:logis}
\end{figure}

In addition to Gaussian bandit and truncated-normal bandit, we also consider logistic bandit with parameter ($\mu = 0, s = 0.5$). The formal definition of logistic distribution and truncated-normal distribution. The results are summarized in Figure \ref{plt:logis}. Giro is almost failed.

\subsection{Linear Bandit.} \label{sec:linear_bandit}

\paragraph{Setup.} We particularly consider the following linear bandit setup. Let $\cD_t\subset \mathbb R^d$ be an arbitrary (finite or infinite) set of arms. When an arm $\bx\in\cD_t$ is pulled, the agent receives a reward
\begin{equation}\label{eqn:reward}
y(\bx) = \bx^{\top}\btheta^* +\epsilon,
\end{equation}
where $\btheta^*\in\mathbb R^d$ is the true reward parameter and $\epsilon$ is a zero-mean random noise with variance $\sigma^2$.
We assume $\|\btheta^*\|_2\leq S$.
An arm $\bx\in\cD_t$ is evaluated according to its expected reward $\bx^{\top}\btheta^*$ and for any $\btheta\in\mathbb R^d$, we denote the optimal arm and its value by 
\begin{equation*}
    \bx^*(\btheta) = \argmin_{\bx\in\cD_t} \bx^{\top}\btheta, \ J(\btheta) = \sup_{\bx\in\cD_t}\bx^{\top}\btheta.
\end{equation*}
Thus $\bx^* = \bx^*(\btheta^*)$ is the optimal arm for $\btheta^*$ and $J(\btheta^*)$ is its optimal value. At each round $t$, the agent selects an arm $\bx_t\in\cD_t$ based on past observations. Then, it observes the reward $y_t = \bx_t^{\top}\btheta^*+\epsilon_t$, and it suffers a regret equal to the difference in expected reward between the optimal arm $\bx^*$ and the arm $\bx_t$. The objective of the agent is to minimize the
cumulative regret up to round $t$, 
\begin{equation*}
R(T) = \sum_{t=1}^T \langle \bx^*-\bx_t, \btheta^*\rangle,
\end{equation*}
where $T$ is the time horizon. Note that the regret holds with high probability and thus is slightly from the standard notion of pseudo regret \citep{abbasi2011improved}.

Denote $\bX_t = (\bx_1,\ldots, \bx_t)^{\top}\in\mathbb R^{t \times d}$, $\by_t = ( y_1,\ldots, y_t)^{\top}\in \mathbb R^{t\times 1}$. At round $t+1$, consider a ridge estimator
	\begin{equation}\label{def:ridge}
	\hat{\btheta}_t = (\bX_t^{\top}\bX_t + \lambda \bI_d)^{-1} \bX_t\by_t.
	\end{equation}
Let us denote $V_t = \sum_{s=1}^t \bx_s\bx_s^{\top}\in\mathbb R^{d\times d}$ as the empirical covariance matrix.

\paragraph{Algorithms.} For TSL: Thompson sampling for linear bandit \citep{agrawal2013thompson}, at each round $t$, the parameter is sampled as $\tilde{\btheta}_t = \hat{\btheta}_t+\hat{\sigma}\sqrt{d\log(1/\delta)}V_t^{-1/2}\eta$ with $\eta\sim \mathbb N(0, I_d)$, where $\hat{\sigma}$ is a standard deviation estimator. \citep{agrawal2013thompson} suggests an even larger constant for the bonus term to enforce over exploration in theory. In practice, it will make the regret exploding. So we remove that large constant in our simulation.

For OFUL: optimism in the face of uncertainty for linear bandits \citep{abbasi2011improved}, at each round $t$, the action is selected as  $\argmax_{\bx}(\bx^{\top}\hat{\btheta}_t + \beta^{\text{OFUL}}_{t,1-\delta, \sigma}\|\bx\|_{V_t^{-1}})$, where 
\begin{equation}
    \beta^{\text{OFUL}}_{t,1-\delta, \sigma} = \hat{\sigma}\sqrt{2\log\Big(\frac{\det(V_t)^{1/2}\det(\lambda \bI_d)^{1/2}}{\delta}\Big) + \lambda^{1/2}S}.
\end{equation}

For BUCBL: bootstrapped UCB for linear bandit, we consider multinomial weights which is equivalent to sample with replacement. In detail, we generate $B$ sets of bootstrap repetitions $\{\bX_t^{(b)}, \by_t^{(b)}\}$ from $\{\bX_t,\by_t\}$ by sample with replacement, and calculate corresponding bootstrapped estimator
	\begin{equation}\label{def:bootstrap_ridge}
	\hat{\btheta}_t^{(b)} = (\bX_t^{(b)\top}\bX_t^{(b)} + \lambda \bI_d)^{-1} \bX_t^{(b)}\by_t^{(b)},
	\end{equation}
and $V_t^{(b)} = \sum_{s=1}^t \bx_s^{(b)}\bx_s^{(b)\top}$. Define the bootstrapped weighted $\ell_2$-norm as follow
\begin{equation*}
    \|\hat{\btheta}_t^{(b)} - \hat{\btheta}_t\|_{V_t^{(b)}+\lambda \bI_d} = \sqrt{(\hat{\btheta}_t^{(b)} - \hat{\btheta}_t)^{\top}(V_t^{(b)} +\lambda \bI_d)(\hat{\btheta}_t^{(b)} - \hat{\btheta}_t )}.
\end{equation*}
For each set of bootstrap repetitions, we could calculate the $ \|\hat{\btheta}_t^{(b)} - \hat{\btheta}_t\|_{V_t^{(b)}+\lambda \bI_d}$ accordingly. Therefore, the bootstrapped threshold is defined as 
\begin{equation}
    q_{\alpha}(\hat{\btheta}_t^{(b)} - \hat{\btheta}_t): = (1-\alpha)\text{-quantile of } \Big\{\|\hat{\btheta}_t^{(1)} - \hat{\btheta}_t\|_{V_t^{(1)}+\lambda \bI_d}, \ldots, \|\hat{\btheta}_t^{(B)} - \hat{\btheta}_t\|_{V_t^{(B)}+\lambda \bI_d}\Big\} .
\end{equation}
At each round $t$, the action is selected as  $\argmax_{\bx}(\bx^{\top}\hat{\btheta}_t +  (q_{\alpha}(\hat{\btheta}_t^{(b)} - \hat{\btheta}_t) + \beta^{\text{OFUL}}_{t,1-\delta, \sigma}/\sqrt{n})\|\bx\|_{V_t^{-1}})$.

\subsection{Logistic Distribution and Truncated-Normal Distribution}\label{subsec:dis}
\paragraph{Logistic Distribution} In probability theory and statistics, the logistic distribution is a continuous probability distribution. Its cumulative distribution function is the logistic function, which appears in logistic regression and feed forward neural networks. It resembles the normal distribution in shape but has heavier tails. 

\begin{definition}
The probability density function (pdf) of the logistic distribution $(\mu, s)$ is given by:
\begin{equation*}
    f(x) = \frac{\exp(-(x-\mu)/s)}{s(1+\exp(-(x-\mu)/s))^2},
\end{equation*}
where $\mu$ is a location parameter and $s>0$ is a scale parameter. The mean is $\mu$ and the variance is $s^2\pi^2/3$.
\end{definition}

\paragraph{Truncated-normal Distribution} In probability and statistics, the truncated normal distribution is the probability distribution derived from that of a normally distributed random variable by bounding the random variable from either below or above (or both). 
\begin{definition}
Suppose $X$ has a normal distribution with mean $\mu$ and variance $\sigma^2$ and lies within the interval $(a, b)$. Then $X$ conditional on $a<X<b$ has a truncated normal distribution $(\mu, a, b)$. Its probability density function $f$ is given by 
\begin{equation*}
    f(x) = \frac{\phi(\frac{x-\mu}{\sigma})}{\sigma(\Phi(\frac{b-\mu}{\sigma}) - \Phi(\frac{a-\mu}{\sigma}))},
\end{equation*}
where $\phi(\cdot)$ is the probability density function of the standard normal distribution and $\Phi(\cdot)$ is its cumulative distribution function.
\end{definition}

%%%%%%%%%%%%%%%%%%%%%%%%%%%%
\section{Supporting Lemmas}\label{sec:supporing_lemma}
%%%%%%%%%%%%%%%%%%%%%%%%%%%

\begin{Lemma}[(Large Deviation Bound, Theorem A.1.4 in \citep{alon2004probabilistic})]\label{lemma:large_deviation}
    Suppose $x_1,\ldots, x_n$ are mutually independent random variables with distribution
\begin{equation*}
     \mathbb P(x_i = 1-p_i) = p_i, \  \mathbb P(x_i = -p_i) = 1-p_i,
\end{equation*}
where $p_i\in[0, 1]$. For any $a>0$, we have 
\begin{equation*}
    \mathbb P\Big(\sum_{i=1}^n x_i >a\Big)< \exp(-2a^2/n).
\end{equation*}

\end{Lemma}

When all $p_i = p$, the sum $\sum_{i=1}^n X_i$ has distribution $\text{Binomial}(n, p)-np$ where $B(n, p)$ is the Binomial distribution.

\begin{Lemma}[(Hoeffding's inequality, Proposition 5.10 in \citep{RV12})]\label{lemma:hoeffding}
    Let $X_1,\ldots, X_n$ be independent centered sub-Gaussian random variables, and let $K = \max_i\|X_i\|_{\phi_2}$. Then for any $\ba = (a_1, \ldots, a_n)^{\top}$ and any $t>0$, we have 
    \begin{equation*}
        \mathbb P\Big(|\sum_{i=1}^na_iX_i|>t\Big)\leq e\exp\Big(-\frac{ct^2}{K^2\|\ba\|_2^2}\Big).
    \end{equation*}
\end{Lemma}

\begin{Lemma}[(Tail Probability for the Sum of Weibull Distributions (Lemma 3.6 in \cite{ALPT2011}))]\label{lemma:alpha_1}
    Let $\alpha\in[1,2]$ and $Y_1,\ldots, Y_n$ be independent symmetric random variables satisfying $\mathbb P(|Y_i|\geq t)=\exp (-t^{\alpha})$. Then for every vector $\ba = (a_1,\ldots, a_n)\in \mathbb R^n$ and every $t\geq 0$,
    \begin{equation*}
        \mathbb P\Big(|\sum_{i=1}^n a_iY_i|\geq t\Big) \leq 2\exp\Big(-c\min\Big(\frac{t^2}{\|\ba\|_2^2}, \frac{t^{\alpha}}{\|\ba\|_{\alpha^*}^{\alpha}}\Big)\Big)
    \end{equation*}
\end{Lemma}

\begin{Lemma}[(Moments for the Sum of Weibull Distributions  (Corollary 1.2 in \cite{bogucki2015suprema}))]\label{lemma:alpha2}
    Let $X_1, X_2,\ldots, X_n$ be a sequence of independent symmetric random variables satisfying $\mathbb P(|Y_i|\geq t)=\exp (-t^{\alpha})$, where $0<\alpha<1$. Then, for $p\geq 2$ and some constant $C(\alpha)$ which depends only on $\alpha$,
    \begin{equation*}
        \left\|\sum_{i=1}^n a_iX_i\right\|_p\leq C(\alpha)(\sqrt{p} \|\ba\|_2+ p^{1/\alpha}\|\ba\|_{\infty}).
    \end{equation*}
\end{Lemma}

\begin{Lemma}[(Khinchin-Kahane Inequality (Theorem 1.3.1 in \cite{de2012decoupling}))]\label{lemma:Khi}
    Let $\{a_i\}_{i=1}^n$ a finite non-random sequence, $\{\varepsilon_i\}_{i=1}^n$ be a sequence of independent Rademacher variables and $1<p<q<\infty$. Then 
    \begin{equation*}
        \Big\|\sum_{i=1}^n \varepsilon_ia_i\Big\|_q \leq \Big(\frac{q-1}{p-1}\Big)^{1/2}\Big\|\sum_{i=1}^n\varepsilon_i a_i\Big\|_p.
    \end{equation*}
\end{Lemma}

\end{document}